\documentclass[10pt,journal,compsoc]{IEEEtran}

\usepackage{cite}
\usepackage{refstyle}
\usepackage{amsmath,amssymb,amsfonts}
\usepackage{algorithm}
\usepackage[noend]{algpseudocode}
\usepackage{textcomp}
\usepackage{xcolor}
\usepackage{booktabs}
\usepackage{array}
\usepackage{multirow}
\usepackage{bm}
\usepackage{hyperref}
\usepackage[noabbrev,capitalize]{cleveref}
\usepackage{makecell}
\usepackage{cuted}
\usepackage{longtable}
\usepackage{url}
\usepackage[english]{babel}
\usepackage{ntheorem}
\usepackage[numbers]{natbib}
\usepackage{graphicx}
\usepackage[labelformat=simple]{subcaption}

\newtheorem{theorem}{Theorem}

\newcommand{\PreserveBackslash}[1]{\let\temp=\\#1\let\\=\temp}
\newcolumntype{C}[1]{>{\PreserveBackslash\centering}p{#1}}
\newcolumntype{R}[1]{>{\PreserveBackslash\raggedleft}p{#1}}
\newcolumntype{L}[1]{>{\PreserveBackslash\raggedright}p{#1}}

\ifCLASSINFOpdf
\else
\fi
\begin{document}
%
    \title{Sharpness-Aware Minimization for Evolutionary Feature Construction in Regression}
%
%
%

    \author{
        Hengzhe Zhang,
        Qi Chen,~\IEEEmembership{Member,~IEEE,}
        Bing Xue,~\IEEEmembership{Fellow,~IEEE,}
        Wolfgang Banzhaf,~\IEEEmembership{Member,~IEEE,}
        Mengjie Zhang,~\IEEEmembership{Fellow,~IEEE}
        \thanks{
            H. Zhang, Q. Chen, B. Xue and M. Zhang are with the School of Engineering and Computer Science, Victoria University of Wellington, PO Box 600, Wellington 6140, New Zealand (e-mails: hengzhe.zhang@ecs.vuw.ac.nz; qi.chen@ecs.vuw.ac.nz; bing.xue@ecs.vuw.ac.nz; and mengjie.zhang@ecs.vuw.ac.nz).\\
            \indent W. Banzhaf is with the Department of Computer Science and Engineering, Michigan State University, East Lansing, MI 48824, USA (e-mails: banzhafw@msu.edu).
        }
    }

%
%

    \markboth{Journal of \LaTeX\ Class Files,~Vol.~14, No.~8, August~2015}%
    {Shell \MakeLowercase{\textit{et al.}}: Bare Demo of IEEEtran.cls for IEEE of Journals}
%



    \IEEEtitleabstractindextext{
        \begin{abstract}
            In recent years, genetic programming (GP)-based evolutionary feature construction has achieved significant success. However, a primary challenge with evolutionary feature construction is its tendency to overfit the training data, resulting in poor generalization on unseen data. In this research, we draw inspiration from PAC-Bayesian theory and propose using sharpness-aware minimization in function space to discover symbolic features that exhibit robust performance within a smooth loss landscape in the semantic space. By optimizing sharpness in conjunction with cross-validation loss, as well as designing a sharpness reduction layer, the proposed method effectively mitigates the overfitting problem of GP, especially when dealing with a limited number of instances or in the presence of label noise. Experimental results on 58 real-world regression datasets show that our approach outperforms standard GP as well as six state-of-the-art complexity measurement methods for GP in controlling overfitting. Furthermore, the ensemble version of GP with sharpness-aware minimization demonstrates superior performance compared to nine fine-tuned machine learning and symbolic regression algorithms, including XGBoost and LightGBM.
        \end{abstract}

    }

    \maketitle

%
    \IEEEpeerreviewmaketitle

    \section{Introduction}
    Automated feature construction is an important task in the domain of automated machine learning (AutoML), which aims to improve the learning performance of a machine learning algorithm on a given dataset $(X, Y)$ by constructing features $\Phi(X)$~\cite{neshatian2012filter}. Automated feature construction methods include neural networks~\cite{lecun2015deep}, kernel methods~\cite{liu2021random}, and genetic programming (GP). Among these, GP has garnered significant attention in recent years due to its interpretable and variable-length representation, as well as its gradient-free search mechanism, making it well-suited for constructing complex and non-differentiable features for non-differentiable base learners and loss functions~\cite{la2018learning}.

    However, the remarkable success of GP in feature construction is counteracted by the challenge of overfitting~\cite{agapitos2019survey}, thereby constraining its broader applicability~\cite{chen2018structural, chen2020rademacher}. Current strategies for mitigating overfitting focus primarily on either the structural risk minimization (SRM) framework or the ensemble learning framework~\cite{nag2015multiobjective}. Examples of SRM-based approaches to combat overfitting include the optimization of Vapnik-Chervonenkis (VC) dimension~\cite{chen2018structural} and Rademacher complexity~\cite{chen2020rademacher}. On the other hand, ensemble learning methods address overfitting by dynamically assembling an ensemble of top GP individuals during the evolutionary process to construct features for decision trees~\cite{zhang2023sr} and sigmoid functions~\cite{wei2020multiclass}.

    However, a recent trend of viewing GP-based feature construction as an efficient evolutionary deep learning technique~\cite{bi2022genetic} prompts us to rethink the techniques for controlling overfitting in GP. In deep learning, the number of parameters often exceeds the number of training instances, resulting in a high VC dimension and Rademacher complexity for the neural network~\cite{zhang2021understanding}. Despite the VC dimension theory suggesting that large deep learning models can memorize and thus overfit the training data~\cite{scarselli1998universal}, these deep models often exhibit robust generalization performance. Therefore, relying on conventional model complexity metrics to control overfitting in feature construction might be too restrictive and insufficient, and alternative approaches should be explored.

    In the theoretical machine learning domain, there exists a theory known as probably approximately correct (PAC)-Bayesian theory that has unveiled that the expected generalization loss, denoted as $\mathbb{E}[\mathcal{L}(f)]$, for a set of deep learning models $f$ with $n$ parameters following a normal distribution $\mathcal{N} (\textbf{w}, \boldsymbol{\sigma})^n$, is bounded as follows with a probability of $1-\delta$~\cite{neyshabur2017exploring}:
    \begin{equation}
        \begin{split}
            \mathbb{E}_{\boldsymbol{\nu} \sim \mathcal{N}(0, \sigma^2)^n}\left[\mathcal{L}\left(f_{\mathbf{w}+\boldsymbol{\nu}}\right)\right]
            \leq
            &\underbrace{\mathbb{E}_{\boldsymbol{\nu} \sim \mathcal{N}(0, \sigma^2)^n}\left[\widehat{\mathcal{L}}\left(f_{\mathbf{w}+\boldsymbol{\nu}}\right)\right]
            -\widehat{\mathcal{L}}\left(f_{\mathbf{w}}\right)}_{\text {Expected Sharpness}}\\
            &+\underbrace{\widehat{\mathcal{L}}\left(f_{\mathbf{w}}\right)}_{\text{Training Loss}}
            +\sqrt{\frac{1}{m}(\underbrace{\frac{\|\mathbf{w}\|_2^2}{2 \sigma^2}}_{\mathrm{KL}}+\ln \frac{2 m}{\delta})}
        \end{split}
        \label{eq: PAC}
    \end{equation}
    In this equation, $\widehat{\mathcal{L}}$ represents the empirical loss function, $\boldsymbol{\nu}$ represents random perturbation parameters following a normal distribution $\mathcal{N} (0, \sigma^2)^n$, and $m$ denotes the number of instances. Intuitively, the first term represents the change in loss after a slight perturbation of parameters, the second term represents the training loss, and the third term represents the deviation of parameters $\mathbf{w}$ from a prior distribution $\mathcal{N} (0, \sigma^2)^n$. Based on this bound, the generalization loss of a machine learning model is partially bounded by the change in loss when the model parameters are slightly perturbed, which is known as expected sharpness.

\begin{figure}[tb]
        \centering
        \begin{subfigure}{0.49\linewidth}
            \includegraphics[width=\linewidth]{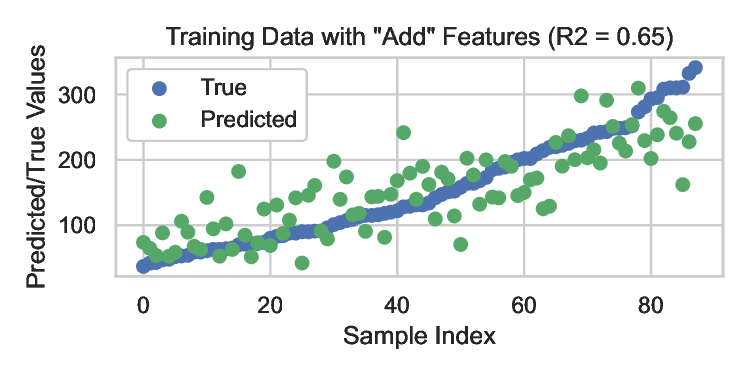}
            \caption{Additive Features}
            \label{fig: Training Add}
        \end{subfigure}
        \begin{subfigure}{0.49\linewidth}
            \includegraphics[width=\linewidth]{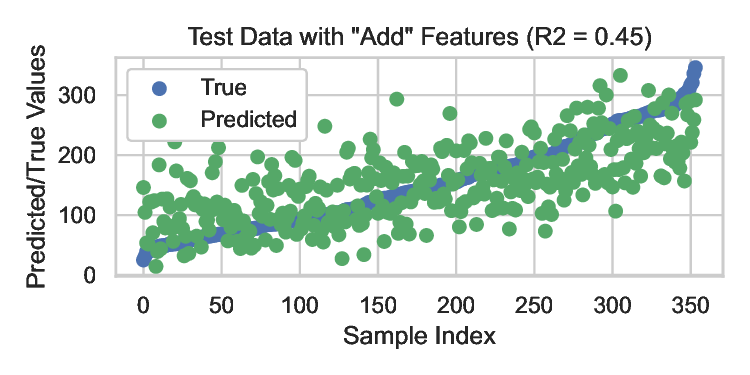}
            \caption{Additive Features}
            \label{fig: Test Add}
        \end{subfigure}

        \vspace{0.05\linewidth}

        \begin{subfigure}{0.49\linewidth}
            \includegraphics[width=\linewidth]{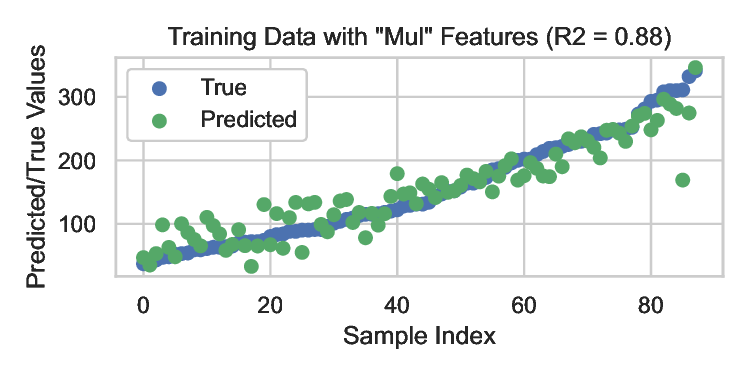}
            \caption{Multiplicative Features}
            \label{fig: Training Mul}
        \end{subfigure}
        \begin{subfigure}{0.49\linewidth}
            \includegraphics[width=\linewidth]{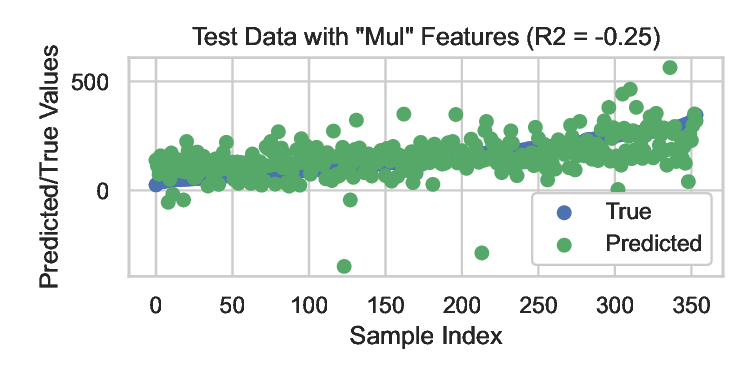}
            \caption{Multiplicative Features}
            \label{fig: Test Mul}
        \end{subfigure}

        \caption{Training and Test $R^2$ Scores for Different Constructed Features on the ``Diabetes'' Dataset.}
        \label{fig: Training-Test}
    \end{figure}

\begin{figure}[tb]
        \centering
        \begin{subfigure}[b]{0.25\linewidth}
            \centering
            \includegraphics[width=\linewidth]{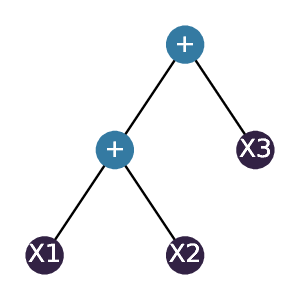}
            \caption{Addition}
            \label{fig: Addition}
        \end{subfigure}%
        \begin{subfigure}[b]{0.3\linewidth}
            \centering
            \includegraphics[width=0.83\linewidth]{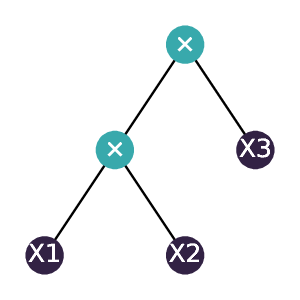}
            \caption{Multiplication}
            \label{fig: Multiplication}
        \end{subfigure}%
        \begin{subfigure}[b]{0.45\linewidth}
            \centering
            \includegraphics[width=\linewidth]{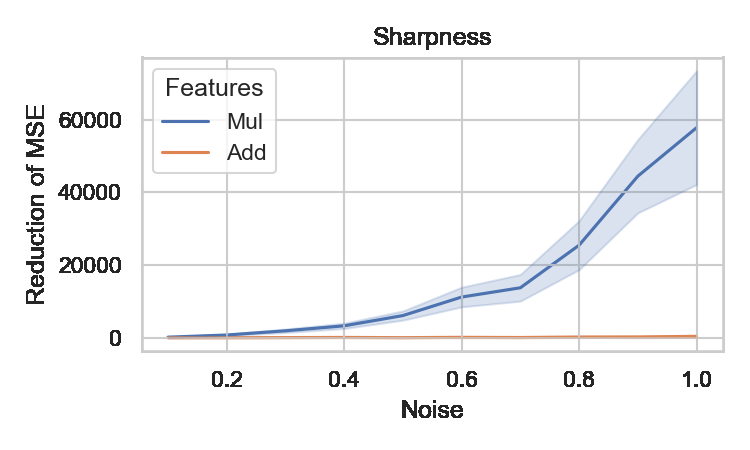}
            \caption{Sharpness}
            \label{fig: sharpness}
        \end{subfigure}

        \caption{Constructed Features and Their Sharpness Values.}

        \label{fig: Sharpness Estimation}
    \end{figure}

    To gain an intuitive understanding of the impact of sharpness minimization on feature construction, i.e., minimizing the first term on the right-hand side of \cref{eq: PAC}, \cref{fig: Training-Test} illustrates an example of the training and test $R^2$ scores using two different sets of constructed features on a dataset named ``Diabetes''~\cite{efron2004least}. One set consists of all possible combinations of third-order addition features, such as $x_1+x_2+x_3$, as shown in \cref{fig: Addition}, while the other comprises all possible third-order multiplication features, like $x_1 \cdot x_2 \cdot x_3$, as depicted in \cref{fig: Multiplication}. In this case, both the number of features and the size of features are identical, indicating that model size cannot effectively distinguish the quality of these two groups of features. Regarding training performance, multiplication features show a better training $R^2$ score than addition features, as shown in \cref{fig: Training Add} and \cref{fig: Training Mul}. However, on the test data, as illustrated in \cref{fig: Test Add} and \cref{fig: Test Mul}, the additive features have a better test $R^2$ score than multiplicative features. This suggests that additive features generalize much better on the test data compared with multiplicative features. In this case, model size cannot effectively control overfitting, but \cref{fig: sharpness} presents promising results of using sharpness as an indicator to detect overfitting. In \cref{fig: sharpness}, we observe that the training loss of additive features is not sensitive to perturbation, whereas the training loss of multiplicative features is highly sensitive to noise/perturbation, implying that multiplicative features correspond to minima of the training loss with high sharpness. Therefore, although multiplicative features achieve higher accuracy on the training data, they are not the preferred choice from the viewpoint of sharpness-aware minimization (SAM) for achieving good generalization performance.

    Based on the above observations, this paper proposes a sharpness-aware minimization framework for GP-based evolutionary feature construction~\footnote{Source Code:\url{https://anonymous.4open.science/r/SAM-GP/experiment/methods/SHARP_GP.py}}. Since GP optimizes the function space rather than the parameter space, this paper proposes a sharpness estimation method to gauge the sharpness of symbolic models. Then, both cross-validation loss and the estimated sharpness are optimized within a multi-objective GP framework to mitigate overfitting in GP. The main objectives of this paper are summarized as follows:
    \begin{itemize}
        \item Introducing a sharpness-aware minimization framework to the symbolic optimization algorithm to address the issue of overfitting in GP-based evolutionary feature construction methods.
        \item Theoretically analyzing the proposed empirical sharpness estimation method for symbolic models, such as GP, within the context of deep learning, considering that GP is a data-efficient deep learning method.
        \item Proposing a sharpness reduction layer in evolutionary feature construction to explicitly minimize sharpness, thereby further enhancing generalization performance on unseen data.
        \item Proposing a bounded prediction strategy for robust predictions, assuming in-distribution predictions, to further enhance generalization performance on unseen data.
        \item Proposing a least recently used caching technique to avoid the repeated perturbation of identical features across different GP individuals, thereby improving the efficiency of sharpness estimation.
        \item Comparing the proposed algorithm with six other methods that optimize different complexity measurements in GP to demonstrate the effectiveness of sharpness-aware minimization, and with nine other machine learning algorithms to demonstrate the effectiveness of GP-based learning methods.
    \end{itemize}

    The remaining sections of this paper are organized as follows. \cref{Related Work} provides an overview of related work on GP for feature construction and overfitting control techniques. \cref{The Proposed Method} presents the details of the proposed method. \cref{Experimental Settings} introduces the experimental settings. \cref{Experimental Results} shows experimental results comparing the proposed method with other complexity methods and other machine learning algorithms. \cref{Further Analysis} offers further analysis regarding sharpness-aware minimization. Finally, in \cref{Conclusion}, we provide conclusions drawn from the study and outline potential directions for future research.

    \section{Related Work}
    \label{Related Work}

    \subsection{Evolutionary Feature Construction}

    Evolutionary feature construction methods have undergone extensive research in recent years. They can be classified into filter-based~\cite{neshatian2012filter}, wrapper-based~\cite{la2020learning, EF-TEVC, chen2020rademacher}, and embedded~\cite{chen2017feature} methods based on the objective function used. In filter-based feature construction methods, the constructed features are evaluated without using any machine learning algorithm but using metrics, such as impurity~\cite{neshatian2012filter}. These approaches offer good time efficiency and generalize well across different machine learning algorithms. However, the constructed features may not achieve top-notch performance on a specific learning algorithm. In contrast, wrapper-based feature construction methods construct features tailored to a specific learning algorithm, thereby achieving good performance on that the wrapped algorithm but at the expense of high computational resources. Finally, embedded methods are as a compromise between filter and wrapper methods, with a typical example being the symbolic regression method~\cite{chen2017feature}, which embeds feature construction in the learning process. In this work, we focus on the wrapper-based feature construction method due to its superior learning performance~\cite{la2020learning, EF-TEVC}.

    \subsection{Controlling Overfitting in Genetic Programming}
    Overfitting is a critical concern within the GP community and has garnered significant attention. Early efforts attempted to control overfitting by managing the model size of GP models, but this approach has limited effectiveness in many cases~\cite{vanneschi2010measuring}. Thus, many methods have been developed to improve the generalization performance of GP.

    \subsubsection{Statistical Learning-based Regularization}
    In recent years, model complexity metrics, such as the degree of the Chebyshev polynomial approximation~\cite{vladislavleva2008order}, VC dimension~\cite{chen2018structural}, and Rademacher complexity~\cite{chen2020rademacher}, have been employed to enhance generalization performance of GP models. These techniques have achieved notable success in controlling functional complexity and improving generalization performance. In addition to model complexity, other approaches from statistical learning theory, such as residual analysis~\cite{chen2020improving}, robust loss function~\cite{raymond2023learning}, and Bayesian model selection~\cite{bomarito2022bayesian}, have also been explored for overfitting control. On the one hand, traditional statistical learning theory often emphasizes model capacity, implying that large/complex models that are capable of memorizing all training data, such as generative pre-trained transformer (GPT), should not generalize well. Specifically, theoretical work has proven that the lower bound of the VC dimension for ReLU-based linear neural networks with $L$ layers and $W$ parameters is $\Omega(WL\log(W/L))$~\cite{bartlett2019nearly}. However, on the other hand, much state-of-the-art work has indicated that large deep learning models can often generalize very well even with high VC dimension and Rademacher complexity that might indicate overfitting of all training data~\cite{jiang2019fantastic}.

    \subsubsection{Implicit Overfitting Control Methods}
    In parallel with the development of modern machine learning techniques, some strategies like random sampling~\cite{gonccalves2013balancing}, semi-supervised learning~\cite{silva2018semi}, soft target~\cite{vanneschi2021soft}, modularization learning~\cite{zhang2023modular}, concept drift detection~\cite{celik2021adaptation} and early stopping~\cite{tuite2011early}, have been employed to mitigate overfitting in GP. While these techniques from the machine learning community are practically useful in some cases, they lack a solid theoretical foundation from generalization theory to explain the effectiveness of these overfitting control techniques.

    \subsubsection{Sharpness-based Regularization}
    Deep representation learning techniques have demonstrated that using a large model for feature construction does not necessarily lead to serious overfitting~\cite{zhang2021understanding}, indicating that the overfitting is more about the behavior of a model, i.e., the semantics of the constructed features. Thus it is necessary to further investigate overfitting control techniques for feature construction. In the deep learning domain, PAC-Bayesian theory-based measures, such as sharpness~\cite{foret2020sharpness}, are more effective in predicting the generalization performance of deep learning models than classical complexity measures~\cite{owen2022standardization}. Thus, it becomes imperative to study overfitting control techniques based on PAC-Bayesian theory and investigate their applicability to GP-based evolutionary feature construction.

    \subsection{Sharpness Aware Minimization}
    The PAC-Bayesian theory has successfully provided practical bounds for the generalization loss of deep neural networks~\cite{neyshabur2017exploring}, sparking a growing interest in utilizing PAC-Bayesian bounds to enhance the generalization performance of deep learning models. Based on the PAC-Bayesian bound presented in \cref{eq: PAC}, the first term $\mathbb{E}_{\boldsymbol{\nu} \sim \mathcal{N}(0, \sigma^2)^n}\left[\widehat{\mathcal{L}}\left(f_{\mathbf{w}+\boldsymbol{\nu}}\right)\right]
    -\widehat{\mathcal{L}}\left(f_{\mathbf{w}}\right)$ has been widely employed to optimize to improve generalization performance, which is known as sharpness-aware minimization (SAM)~\cite{foret2020sharpness}. The underlying intuition is to find a model situated at the flat minima of the loss landscape, leading to a lower expected generalization loss. The SAM process generally comprises two main steps:
    \begin{itemize}
        \item Sharpness Maximization: Gradient ascent or random search is applied to perturb the weights of a model $w$ within a certain radius $\rho$, aiming to find the worst point of training loss in the neighborhood of the current model.
        \item Loss Minimization: The gradient obtained from the perturbed model is utilized to update the weights of the original model, guiding the model toward a flat minimum.
    \end{itemize}
    To achieve sharpness-aware minimization effectively and efficiently in real-world learning algorithms, several aspects have been taken into account and studied:
    \begin{itemize}
        \item Optimization Direction: Depending on whether the sharpness maximization direction is computed based on the entire batch size $n$ or a subset of $m$ data samples, the approaches are categorized as $n$-sharpness and $m$-sharpness. In practice, $m$-sharpness is more effective~\cite{andriushchenko2022towards}. $m$-sharpness has a special case that focuses on maximizing sharpness around a neighborhood $\rho$ independently for each data point, i.e., $m=1$. In this case, optimizing $m$-sharpness can be equivalent to optimizing average sharpness in PAC-Bayesian bounds when $\rho$ and the learning rate are sufficiently small~\cite{wen2022sharpness}. Moreover, during optimization, the gradient of the original model may conflict with the gradient of the perturbed model. Projecting the original gradient onto the perturbed gradient can yield better results than optimizing either one alone~\cite{zhuang2021surrogate}.
        \item Optimization Step Size: As weights may vary in scale, using adaptive perturbation based on weight scale can enhance performance~\cite{kwon2021asam}.
        \item Efficient Perturbation: Even though estimating sharpness with gradient ascent is computationally intensive, one-step gradient ascent can serve as an approximation. To further reduce computational costs, sharpness can also be estimated through random perturbation. This approach is effective if the randomly perturbed weights are simultaneously optimized in parallel with the original weights~\cite{li2022efficient}.
    \end{itemize}
    While sharpness-aware minimization has made significant advancements in the domain of deep learning~\cite{foret2020sharpness}, it has not yet been explored in the context of GP-based evolutionary feature construction. This is primarily because GP optimizes the function space rather than the parameter space. Therefore, GP lacks a clear definition of weights, and thus it is necessary to develop a method for sharpness estimation that is suitable for GP, which is semantic perturbation in this paper.

    \section{The Proposed Method}
    \label{The Proposed Method}
    In this section, we introduce a framework for sharpness-aware minimization in GP-based evolutionary feature construction. The section begins with an overview of the sharpness-aware minimization framework for feature construction. Subsequently, we present a practical method for estimating sharpness in GP, supported by theoretical analysis. To further reduce the sharpness of constructed features, a sharpness reduction layer is proposed. Finally, we propose two strategies, namely bounded prediction and ensemble learning, to enhance generalization performance further.

    \subsection{Algorithm Framework}
    \label{tab: Framework}
\begin{figure}[tb]
        \centering
        \includegraphics[width=\linewidth]{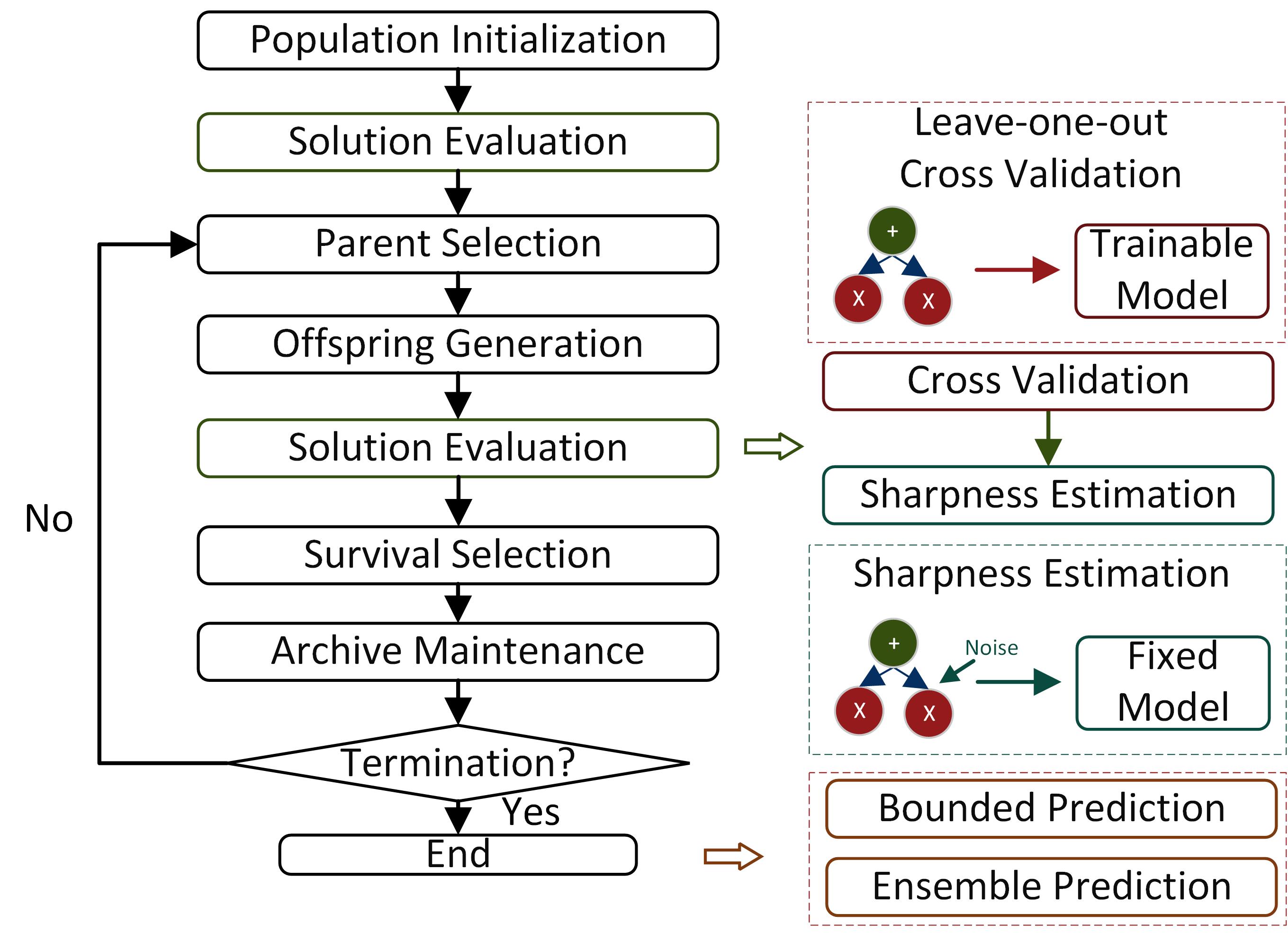}
        \caption{Algorithm Framework}
        \label{fig: Algorithm Framework}
    \end{figure}

    The sharpness-aware feature construction algorithm proposed in this paper adheres to the conventional framework of evolutionary feature construction algorithms, as illustrated in \cref{fig: Algorithm Framework}, and includes the following steps:
    \begin{itemize}
        \item Initialization: Initially, $N$ individuals are generated randomly. Each individual includes one GP tree initialized using the ramped half-and-half method~\cite{banzhaf1998genetic}, representing a constructed feature.
        \item Solution Evaluation: This phase consists of three steps.
        \begin{itemize}
            \item Feature Construction: First, features $\Phi(X)$ are constructed based on the GP trees $\Phi$.
            \item Cross-Validation: Next, a linear model with L2 regularization is trained using these constructed features $\Phi(X)$ to make predictions and calculate the mean squared error on the training data, serving as the first objective $O_1(\Phi)$. The leave-one-out cross-validation (LOOCV) scheme is employed on the training set to mitigate overfitting. Specifically, for a regularization strength $\alpha$, the weight $\boldsymbol{w}$ and leverage $\boldsymbol{h}$ are first calculated as shown in \cref{eq: weight and leverage}:
            \begin{equation}
                \begin{split}
                    \boldsymbol{w} &= \left( \mathbf{X}^\top \mathbf{X} + \alpha \mathbf{I} \right)^{-1} \mathbf{X}^\top \mathbf{Y},\\
                    \boldsymbol{h} &= \text{diag}\left(\mathbf{X} \left( \mathbf{X}^\top \mathbf{X} + \alpha \mathbf{I} \right)^{-1} \mathbf{X}^\top\right).
                \end{split}
                \label{eq: weight and leverage}
            \end{equation}
            Then, the leave-one-out cross-validation errors on training set $\boldsymbol{e}$ are computed based on the weight and leverage as shown in \cref{eq: LOOCV}:
            \begin{equation}
                \boldsymbol{e} = \left( \frac{\mathbf{Y} - \mathbf{X} \boldsymbol{w}}{1 - \boldsymbol{h}} \right)^2.
                \label{eq: LOOCV}
            \end{equation}
            \item Sharpness Estimation: Finally, the sharpness of the model is estimated, serving as the second objective value $O_2(\Phi)$. Specifically, constructed features are perturbed, and predictions are made based on perturbed features to estimate sharpness. The method for estimating sharpness will be described in \cref{Empirical Sharpness Estimation}.
        \end{itemize}
        \item Parent Selection: For a population $P$, parent individuals are selected based on lexicase selection. Lexicase selection is a semantic-based selection operator focusing on both improving accuracy and diversity, but does not incorporate the sharpness objective. The sharpness objective is mainly optimized through the survival selection operator after the offspring generation stage. The core idea of lexicase selection is to iteratively construct filters to eliminate individuals based on the median absolute deviation of the loss $\epsilon_{k}$ for a randomly selected training instance $k$, i.e., only keeping individuals $\Phi$ that satisfy $\mathcal{L}_k(\Phi) \leq \min_{\Phi' \in P} \mathcal{L}_k(\Phi')+\epsilon_{k}$, where $\min_{\Phi' \in P} \mathcal{L}_k(\Phi')$ represents the best error on the instance. This process is repeated until only one individual remains.
        \item Offspring Generation: After parent individuals are selected, offspring are generated by applying random subtree crossover and random subtree mutation operators. Moreover, we also employ a tree addition operator and a tree deletion operator to support adding and deleting features~\cite{la2020learning}, allowing GP to automatically determine the appropriate number of trees/features. Specifically, the tree addition operator randomly generates a tree and adds it to an individual, whereas the feature tree operator randomly selects a tree and deletes it from an individual.
        \item Survival Selection: Based on the two objectives, environmental selection operators from NSGA-II~\cite{deb2002fast} are used to select $|P|$ individuals from a combination of $|P|$ parents and $|P|$ offspring. Briefly, the dominance relationship and crowding distance between two selected individuals are sequentially checked to determine the superior individual. Formally, an individual $\Phi_1$ dominates another individual $\Phi_2$ if the following two conditions are satisfied:
        \begin{equation}
            \begin{split}
                \forall i \in \{1, 2\}, O_i(\Phi_1) \leq O_i(\Phi_2)\\
                \exists j \in \{1, 2\}, O_j(\Phi_1) < O_j(\Phi_2)
            \end{split}
        \end{equation}
        The crowding distance is defined as the average distance of a point to its two nearest neighbors in the objective space. If a superior individual cannot be identified using these criteria, an individual is randomly selected as the parent.
        \item Archive Maintenance: At this stage, the historically best-performing individual based on the sum of cross-validation loss and the sharpness value, i.e., $O_1(\Phi)+O_2(\Phi)$, is archived. Typically, only one individual is stored, which is subsequently used for making predictions on unseen data. However, more individuals could be stored in the archive to make an ensemble prediction on unseen data, but an ensemble model may reduce interpretability.
    \end{itemize}

    \subsection{Sharpness Aware Minimization (SAM)}
    For any feature construction method, including neural networks and GP, sharpness is defined as the increase in loss after a slight perturbation of features $\Phi$. Assuming the perturbed features are denoted by $\tilde{\Phi}$, the objective functions for SAM of a learning model $f$ are as follows:
    \begin{equation}
        \min_\Phi (\underbrace{\widehat{\mathcal{L}}(f(\Phi(X)),Y)}_{\text{Training Loss}}, \underbrace{\mathbb{E}_{\tilde{\Phi}}\; \widehat{\mathcal{L}}(f(\tilde{\Phi}(X)),Y)-\widehat{\mathcal{L}}(f(\Phi(X)),Y)}_{\text{Expected Sharpness}})
    \end{equation}
    The first term is the objective traditionally optimized, while the second term measures the sharpness of a set of features $\Phi$. For features $\Phi$ constructed by a neural network with a set of parameters $w$, sharpness is often approximated using the one-step gradient ascent method~\cite{foret2020sharpness}. Therefore, sharpness is defined as $\widehat{\mathcal{L}}\left(f(w+\rho \frac{\nabla_w L(x)}{\|\nabla_w L(x)\|_2},X), Y\right)-\widehat{\mathcal{L}}\left(f(w,X), Y\right)$, where $\rho$ denotes the perturbation range, and $\frac{\nabla_w L(x)}{\|\nabla_w L(x)\|_2}$ is the normalized gradient used to determine the direction of perturbation. Thus, the empirical objective functions become:

    \begin{equation}
        \text { minimize }
        \left\{
        \begin{array}{l}
            O_1(\Phi)=\widehat{\mathcal{L}}\left(f(w,X), Y\right) \\
            O_2(\Phi)=\widehat{\mathcal{L}}\left(f(w+\rho \frac{\nabla_w L(x)}{\|\nabla_w L(x)\|_2},X), Y\right)-O_1(\Phi)
        \end{array}
        \right.
    \end{equation}
    Here, $O_1(\Phi)$ is the empirical loss and $O_2(\Phi)$ is sharpness. Regarding the perturbation term $\rho \frac{\nabla_w L(x)}{|\nabla_w L(x)|_2}$, it can be calculated for each instance or computed based on a batch of data, resulting in different schemes known as 1-sharpness and n-sharpness. The idea of 1-sharpness or n-sharpness can be applied to optimization in function space. However, calculating the gradient $\rho \frac{\nabla_w L(x)}{\|\nabla_w L(x)\|_2}$ for GP-based feature construction is challenging due to the unclear definition of learnable weights $w$ in function space. Therefore, we propose a method to estimate sharpness in \cref{Empirical Sharpness Estimation}.

    \subsection{Empirical Sharpness Estimation}
    In this section, we introduce an empirical method to estimate sharpness in the context of GP. The section begins by describing how to perturb the semantics/outputs of GP trees to explore the neighborhood of GP trees in the semantic space. Then, based on perturbed GP trees from \cref{sec: Semantic Perturbation}, the method for calculating the worst sharpness of a GP individual is introduced in \cref{sec: Semantic Sharpness Estimation}. This paper employs a 1-SAM strategy instead of an n-SAM strategy to calculate the worst sharpness, and their differences in GP are introduced in \cref{sec: 1-SAM}. The relationship between sharpness estimation in GP and neural networks is illustrated in \cref{sec: SAM for GP vs. NN}.

    \label{Empirical Sharpness Estimation}

    \subsubsection{Semantic Perturbation}
    \label{sec: Semantic Perturbation}
    In contrast to NN, GP primarily focuses on optimizing the structure $\Phi$ in function space rather than the weights $w$ in parameter space, making it challenging to apply perturbations directly. To address this issue, we introduce a method for perturbing the inputs of each non-terminal node $\psi_c$, instead of perturbing the structure, i.e., perturbing the semantics/outputs of each subtree of a GP tree, including terminal nodes. Formally, suppose the semantics/outputs of a subtree in a GP tree $a$ are $\psi_{a}(X)$ for input data $X$; then, the perturbed outputs $\tilde{\psi}_{a}(X)$ are defined as:
    \begin{equation}
        \tilde{\psi}_{a}(X) = \psi_{a}(X) + \psi_{a}(X) \cdot \mathcal{N}(0, \sigma^2)
        \label{label: ouput}
    \end{equation}
    where $\mathcal{N}(0, \sigma^2)$ represents Gaussian noise generated with a mean of zero and a standard deviation of $\sigma$. The Gaussian noise is scaled by the semantics $\psi_{a}(X)$ to adjust the magnitude of the perturbation to match that of the semantics. For a parent node $\text{FunctionNode}_{c}$ in a GP tree that receives outputs from nodes $a$ and $b$ as inputs, the semantics of the GP tree $\psi_{c}$, based on a single node $c$ and its perturbed subtree $\tilde{\psi}_{a}$ and $\tilde{\psi}_{b}$, are defined by:
    \begin{equation}
        \psi_{c}(X) = \text{FunctionNode}_{c}(\tilde{\psi}_{a}(X), \tilde{\psi}_{b}(X)).
        \label{label: input}
    \end{equation}
    This equation illustrates perturbation on a single node. In a typical GP tree with recursive subtrees, the perturbation process is recursively applied to each subtree $\psi$ of a GP tree $\phi$, resulting in a perturbed feature $\tilde{\phi}(X)$. All perturbed features constitute a perturbed individual $\tilde{\Phi}(X)$.

    \subsubsection{Semantic Sharpness Estimation}
    \label{sec: Semantic Sharpness Estimation}
    The entire process for sharpness estimation is provided in \cref{alg: SAM}, which consists of four stages:
    \begin{itemize}
        \item Original Prediction (Lines 2-3): In the initial stage of sharpness estimation, we obtain features constructed by GP trees, denoted as $\Phi(X)$, and the corresponding predictions $\hat{Y}$ based on the trained linear model $LM$, i.e., $\hat{Y}_i=LM(\Phi(X))$.
        \item Cache-based Random Perturbation (Line 5): During each iteration of the perturbation process, Gaussian noise $\mathcal{N} (0, \sigma^2)$ is added to the semantics of each subtree $\psi$ within each GP tree $\phi$ in the GP individual $\Phi$, according to \cref{label: ouput} and \cref{label: input}. The Gaussian noise is controlled by the random seed $k \in K$. For this random perturbation process, a least recently used cache is constructed for each random seed $k \in K$. If a GP tree $\phi$ has been randomly perturbed in other GP individuals, the perturbed feature $\tilde{\Phi}(X)$ is directly retrieved from the cache to save computational resources.
        \item Perturbed Prediction (Line 6): Once perturbed features $\tilde{\Phi}(X)$ are generated, these features are fed into the trained linear model $LM$ to produce perturbed predictions $\tilde{Y}$, i.e., $\tilde{Y}_i=LM(\tilde{\Phi}(X))$.
        \item Sharpness Calculation (Lines 7-8): Finally, sharpness is calculated as the difference between the mean squared errors of $\tilde{Y}$ and $\hat{Y}$, specifically $(\tilde{Y}-Y)^2-(\hat{Y}-Y)^2$. Here, we use $1$-sharpness, and thus, the worst mean squared error on one instance/fitness case is recorded as $S_i$ separately.
    \end{itemize}
    To ensure the optimization of the worst-case sharpness~\cite{andriushchenko2022towards} rather than the average sharpness, random perturbations and sharpness calculations are repeated $K$ times on each GP tree. The final sharpness value $S_i$ for each instance $i$ is the maximum sharpness over $K$ iterations, i.e., $S_i=\max_{k=1}^{K} (\tilde{Y_{i,k}} - Y_i)^2 - (\hat{Y_i} - Y_i)^2$ across the $k$ rounds of perturbed predictions $\tilde{Y_{i,k}}$ for instance $i$. For a GP individual, the sharpness value is the average sharpness value across $N$ instances, i.e., $\sum_{i=1}^{N} S_i /N$. In summary, the objective function of sharpness-aware minimization in GP is:
    \begin{equation}
        \text { minimize }
        \left\{
        \begin{array}{l}
            O_1(\Phi)=\sum_{i=1}^{N} (\hat{Y_i}-Y_i)^2 / N \\
            O_2(\Phi)=\sum_{i=1}^{N} S_i / N
        \end{array}
        \right.
    \end{equation}
    where $O_1(\Phi)$ represents the empirical mean squared error, and $O_2(\Phi)$ is the estimated sharpness. In practice, $O_1(\Phi)$ is replaced by the cross-validation error, as provided in \cref{eq: LOOCV}.
    \begin{table*}[tb]
        \centering
        \caption{Strategies in Sharpness-Aware Minimization for Neural Networks and Evolutionary Feature Construction. All these strategies are integrated in this paper.}
        \begin{tabular}{ccc}%
            \toprule
            \textbf{Strategies}                                      & \textbf{Neural Network}                   & \textbf{Evolutionary Feature Construction}  \\%
            \midrule
            Sharpness Estimation~\cite{foret2020sharpness}           & Gradient Ascent                           & Random Search                               \\
            $m$-Sharpness~\cite{andriushchenko2022towards}           & Mini-batch Sharpness                      & Mini-batch Sharpness                        \\
            Random Perturbation~\cite{li2022efficient}               & Random Perturbation on Weights            & Random Perturbation on Semantics            \\
            Worst-case Perturbation~\cite{andriushchenko2022towards} & Gradient Ascent                           & Maximum of multiple rounds Estimation       \\
            Adaptive Sharpness~\cite{kwon2021asam}                   & Perturbation Scaling According to Weights & Perturbation Scaling According to Semantics \\
            \bottomrule
        \end{tabular}%
        \label{tab: sharpness estimation}%
    \end{table*}%

    \subsubsection{1-SAM for GP}
    \label{sec: 1-SAM}
\begin{figure}[tb]
        \centering
        \includegraphics[width=0.65\linewidth]{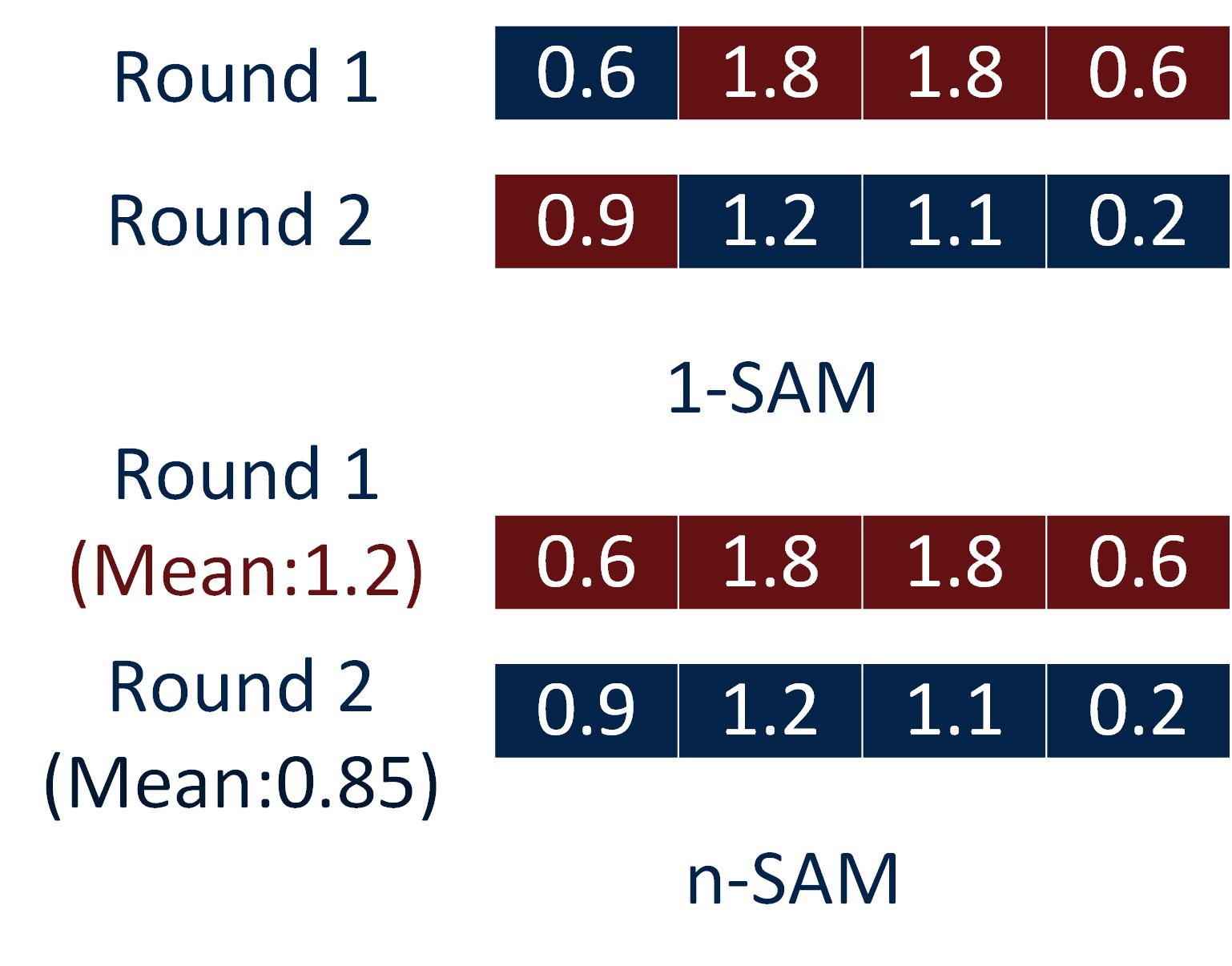}
        \caption{Differences between 1-SAM and n-SAM. Values in the figure denote sharpness for each training instance, with red blocks representing the final selected sharpness values and blue blocks representing unused sharpness values.}
        \label{fig: 1-SAM}
    \end{figure}

    The sharpness estimation process for GP introduced in this paper is 1-SAM. An example illustrating the difference between 1-SAM and n-SAM is shown in \cref{fig: 1-SAM}. In the case of 1-SAM, the final sharpness is calculated based on the maximum sharpness value $S_i$ within each instance $i \in \{1, \dots, N\}$ across $K$ rounds. Thus, the overall sharpness $S_{\text{1-SAM}}$ is defined as:
    \begin{equation}
        S_{\text{1-SAM}} = \frac{1}{N} \sum_{i = 1, \dots, N} \max_{k = 1, \dots, K} (\tilde{Y_i^k} - Y_i)^2 - (\hat{Y_i} - Y_i)^2.
    \end{equation}
    The mean of all the maximum sharpness values across all instances is computed as the final sharpness, i.e., $\sum_{i=1}^N S_i$. In contrast, for n-SAM, sharpness is calculated by perturbing the entire batch of data all at once, rather than one by one. If there are $K$ rounds of perturbation, the largest mean sharpness over the $K$ rounds is used as the final sharpness. In this case, the overall sharpness $S_{\text{n-SAM}}$ is defined as:
    \begin{equation}
        S_{\text{n-SAM}} =  \max_{k = 1, \dots, K} \frac{1}{N} \sum_{i = 1, \dots, N} (\tilde{Y_i^k} - Y_i)^2 - (\hat{Y_i} - Y_i)^2.
    \end{equation}
    In this work, our goal is to optimize 1-sharpness instead of n-sharpness because it has been shown to have superior performance in controlling overfitting in deep learning~\cite{foret2020sharpness}.

    \subsubsection{SAM for GP vs. NN}
    \label{sec: SAM for GP vs. NN}
\begin{figure}[tb]
        \centering
        \includegraphics[width=0.7\linewidth]{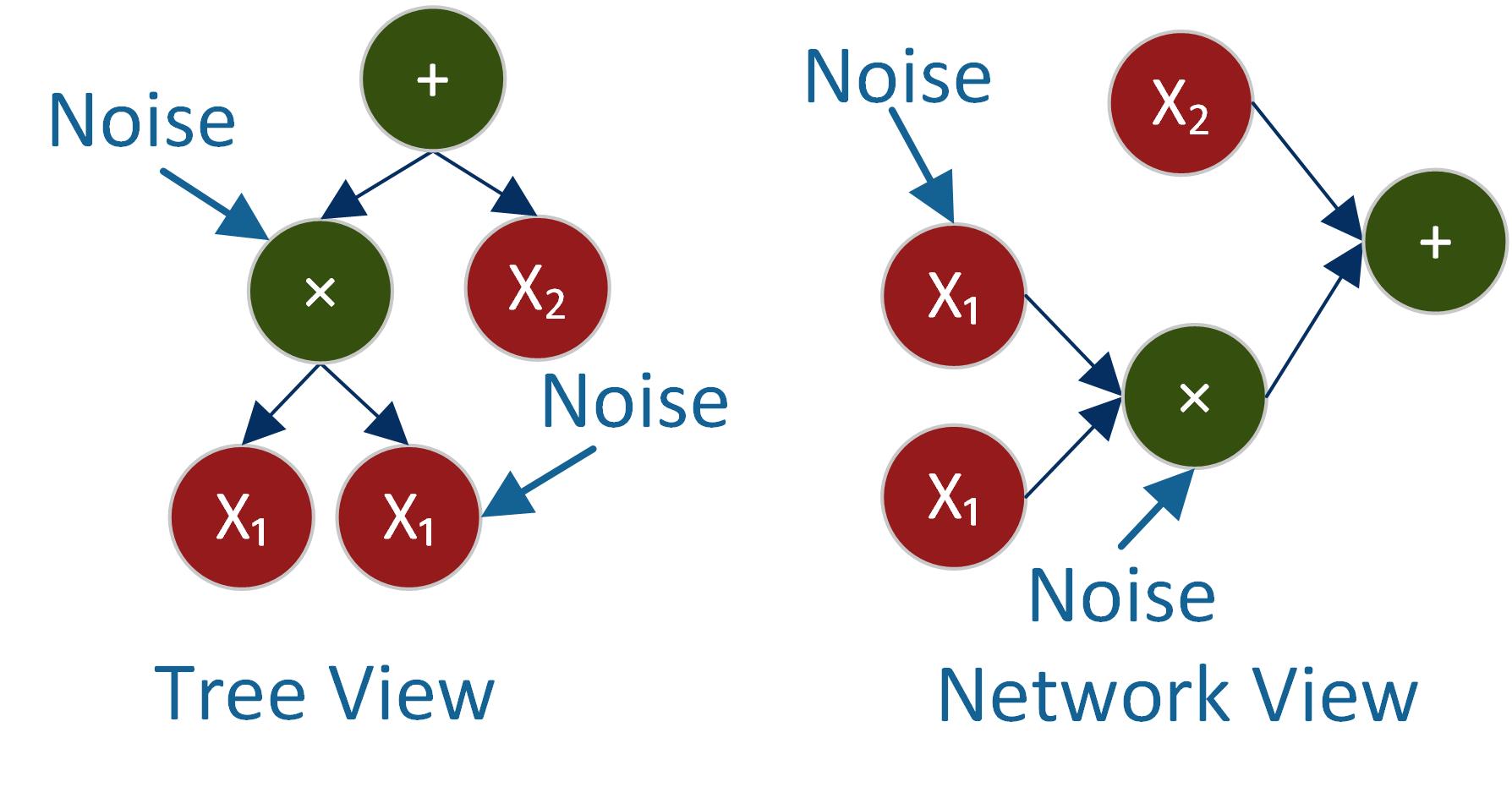}
        \caption{Two perspectives on sharpness estimation for GP: GP can be considered an efficient deep learning technique~\cite{la2018learning,gaier2019weight} when the GP tree is rotated 90 degrees.}
        \label{fig: Sharpness}
    \end{figure}

    \cref{tab: sharpness estimation} provides a summary of the differences in strategies used in sharpness estimation in NN and GP-based evolutionary feature construction. The symbolic nature of GP and its non-differentiable properties lead to distinctions in the sharpness estimation method. However, there is still a connection between sharpness estimation in GP and NN. \cref{fig: Sharpness} offers a dual perspective on the sharpness estimation process proposed in this paper. From the viewpoint of a GP tree, the sharpness estimation process perturbs the semantics of subtrees within the tree. From the neural network perspective, the process perturbs the inputs of each layer. In PAC-Bayesian theory~\cite{neyshabur2017exploring}, sharpness is defined as the change in loss resulting from parameter perturbations. To clarify the relationship between adding noise to weights and inputs at each layer, theoretical analysis is presented in \cref{Theoretical Analysis}. This analysis demonstrates that in the case of a neural network, adding noise to the inputs at each layer is equivalent to perturbing the weights. Thus, in the context of GP, since it is challenging to define weight perturbation in function space, this can be achieved by perturbing the inputs of each subtree in GP, i.e., changing the semantics/outputs of all the children of each subtree, rather than perturbing the weights.

    \begin{algorithm}[!tb]
        \caption{Semantic Sharpness Estimation for GP Trees.}
        \label{alg: SAM}
        \begin{algorithmic}[1]
            \Require GP Tree $\Phi$, Inputs $X$, Target Outputs $Y$, Linear Model $LM$, Gaussian Noise Standard Deviation $\sigma$, Number of Iterations $K$
            \State Initialize sharpness $S \gets 0$
            \State $\Phi(X) \gets $ Feature Construction ($X, \Phi$)
            \State $\hat{Y} \gets$  Prediction($LM, \Phi(X)$)
            \For{$k = 1, \ldots, K$}
                \State $\tilde{\Phi}(X) \gets $ Feature Construction with Noise ($X, \Phi, \sigma$)
                \State $\tilde{Y^k} \gets$  Prediction($LM, \tilde{\Phi}(X)$)
                \For{$i = 1, \ldots, N$}
                    \State $S_i \gets \max({S_i , (\tilde{Y_i^k} - Y_i)^2 - (\hat{Y_i} - Y_i)^2})$
                \EndFor
            \EndFor
            \Ensure Sharpness Value $\sum_{i=1}^{N} S_i /N$
        \end{algorithmic}
    \end{algorithm}

    \subsection{Theoretical Foundation of Sharpness Estimation}
    \label{Theoretical Analysis}

    \begin{theorem}
        For a layer in a neural network represented as $f(x)=g(\sum_{i=1}^{k} w_i x_i)$ with an arbitrary activation function $g$, adding adaptive noise $\mathcal{N}(0, w_i^2)^k$~\cite{neyshabur2017exploring, kwon2021asam} to weights $w$ is equivalent to adding noise of $\mathcal{N}(0, x_i^2)^k$ to the inputs of this layer, where $k$ denotes the dimensionality of the input data for this layer, and $x_i$ represents the value for each dimension $i$ of input $x$.
    \end{theorem}

    \textbf{Proof:}
    Through mathematical transformation, the inputs of the activation function, $\sum_{i=1}^{k} w_i x_i$, can be transformed as follows:

    \begin{equation}
        \label{tab: proof}
        \begin{split}
            \sum_{i=1}^{k} (w_i+\epsilon_i)x_i
            &=\sum_{i=1}^{k} w_i(1+\epsilon_i/w_i)x_i\\
            &=\sum_{i=1}^{k} w_i(x_i+x_i \epsilon_i/w_i)
        \end{split}
    \end{equation}

    Given that $x_i$ and $w_i$ are constants when adding noise, let the noise added to $x_i$ be denoted as $\theta_i$, i.e., $\theta_i=\epsilon_i w_i/x_i$. Then, $\theta_i$ follows the following cumulative distribution:
    \begin{equation}
        \begin{split}
            CDF(x_i \epsilon_i/w_i)&=P(x_i \epsilon_i/w_i<k)\\
            &=P(\epsilon_i<k w_i/x_i)\\
            &=\int_{-\infty}^{k w_i/x_i} \frac{1}{w_i \sqrt{2 \pi}} e^{-\frac{1}{2}\left(\frac{\epsilon_i}{w_i}\right)^2} d\epsilon_i\\
            &=\int_{-\infty}^{k} \frac{1}{w_i \sqrt{2 \pi}} e^{-\frac{1}{2}\left(\frac{w_i \theta_i}{w_i x_i}\right)^2} \frac{w_i}{x_i} d\theta_i\\
            &=\int_{-\infty}^{k} \frac{1}{x_i \sqrt{2 \pi}} e^{-\frac{1}{2}\left(\frac{\theta_i}{x_i}\right)^2} d\theta_i\\
        \end{split}
    \end{equation}
    From the cumulative distribution, it is clear that $\theta_i=x_i \epsilon_i/w_i$ follows a Gaussian distribution $\mathcal{N}(0, x_i^2)$. Therefore, adding noise to the inputs in each dimension is equivalent to adding adaptive noise to the weight of each dimension. Furthermore, since $k$ random noises are added to $k$ dimensions, and $\epsilon_i, \ldots, \epsilon_k$, are independently sampled, the conclusion in one dimension can be extended to all dimensions. As a result, Theorem 1 has been proven. An empirical verification is provided in the supplementary material. Now, given that perturbing parameters is equivalent to perturbing inputs, and considering that GP does not have a clear definition of weights, it is reasonable to perturb the inputs of each layer in GP, i.e., the semantics/outputs of the previous layer of each layer.

    \subsection{Sharpness Reduction Layer}
    The conventional GP operators, including mathematical operators such as $\times$ and $\div$, are highly sensitive to even minor changes in input, which can lead to altered computation results and result in non-zero sharpness. To address this issue, we propose a sharpness reduction layer, the key idea of which is to denoise corrupted results. During the training process, the value of $ \psi^*_{c}(X)=\text{FunctionNode}_{c}(\psi_{a}(X), \psi_{b}(X))$ is stored alongside node $c$. In the inference stage, a binary search is applied to the pre-sorted $ \psi_{c}(X)$ to search for the nearest value, $\psi_c^+(x)$, that matches the subtree value $\psi_c(x)$ on the test instance $x$, as outlined in \cref{alg: Sharpness Reduction Layer}. Then, the subtree value on the test instance $x$, $\psi_c(x)$, is replaced by the nearest value $\psi_c^+(x)$. For a GP tree with multiple layers, the sharpness reduction layer is applied to each layer to stabilize the semantics of constructed features. The sharpness reduction layer avoids abrupt changes in constructed features in the neighborhood of training samples, which could lead to sharp increases in training loss. By applying this sharpness reduction layer, the final model could be made more robust to unseen data and have improved generalization performance.

    \begin{algorithm}[!tb]
        \caption{Sharpness Reduction Layer}
        \label{alg: Sharpness Reduction Layer}
        \begin{algorithmic}[1]
            \Require Training Data $\psi_c(X)$, Test Data $\psi_c(x)$
            \State $\text{index} \gets \text{Binary Search}(\psi_c(X), \psi_c(x))$
            \State $\text{index} \gets \text{clip}(\text{index}, 1, \text{len}(X) - 1)$
            \Comment{Handle boundary values}
            \State $\psi_c(x_{\text{left}}) \gets \psi_c(X)[\text{index} - 1]$
            \State $\psi_c(x_{\text{right}}) \gets \psi_c(X)[\text{index}]$
            \If{$(\psi_c^+(x) - \psi_c(x_{\text{left}})) \leq (\psi_c(x_{\text{right}}) - \psi_c^+(x))$}
                \State $\psi_c^+(x) \gets \psi_c(x_{\text{left}})$
            \Else
                \State $\psi_c^+(x) \gets \psi_c(x_{\text{right}})$
            \EndIf
            \Ensure $\psi_c^+(x)$
        \end{algorithmic}
    \end{algorithm}

    \subsection{Post-Processing Strategies}
    To further enhance the generalization performance of GP-based evolutionary feature construction, we propose two strategies: bounded prediction and ensemble learning.

    \subsubsection{Bounded Prediction}
    In machine learning, it is usually assumed that unseen data are generated from the same distribution as the training data. Consequently, predictions should not deviate significantly from the labels in the training data. To account for this, during training, we record both the maximum target values $\max_{i=1}^N Y_i$ and the minimum target values $\min_{i=1}^N Y_i$ from the training dataset. Then, when making predictions on unseen data, we constrain the prediction $\hat{Y_i}$ within the range defined by $\min_{i=1}^N Y_i$ and $\max_{i=1}^N Y_i$. This constraint is commonly found in many popular machine learning algorithms, such as decision trees and k-nearest neighbors. Formally, the bounded prediction $\hat{Y}_{\text{bounded}}$ is defined as:
    \begin{equation}
        \hat{Y}_{\text{bounded}} =
        \begin{cases}
            \min_{i=1}^N Y_i & \text{if } \hat{Y} < \min_{i=1}^N Y_i \\
            \max_{i=1}^N Y_i & \text{if } \hat{Y} > \max_{i=1}^N Y_i \\
            \hat{Y} & \text{otherwise}
        \end{cases}
    \end{equation}
    By leveraging the assumption that the data are generated from the same distribution, we anticipate that this bounded prediction step can enhance generalization performance of the proposed method.

    \subsubsection{Ensemble Learning}
    Ensemble learning is an effective technique for reducing algorithm variance, thereby improving generalization performance on unseen data. Given that GP is a population-based optimization algorithm, it is natural to maintain an archive $A$ of the top $|A|$ individuals based on the sum of the average leave-one-out cross-validation loss and sharpness values during the evolutionary process to form the ensemble. In other words, during the archive maintenance stage in \cref{tab: Framework}, not only the best individual is saved but also the top $A$ individuals. Then, ensemble predictions are generated by averaging the predictions from all models $\hat{Y}_{\Phi}$ within the ensemble.
    Formally, the ensemble prediction $\hat{Y}_{\text{ensemble}}$ is defined as follows:
    \begin{equation}
        \hat{Y}_{\text{ensemble}}=\frac{1}{|A|} \sum_{\Phi \in A} \hat{Y}_{\Phi}
    \end{equation}
    It is worth noting that ensemble learning is an optional component for SAM-based GP that can further enhance generalization performance. It is not recommended when interpretability is the primary focus.

    \section{Experimental Settings}
    \label{Experimental Settings}
    This section presents the experimental settings used to compare SAM and other overfitting control techniques for GP-based feature construction, including benchmark datasets, parameter settings, evaluation protocols, and baseline methods.

    \subsection{Benchmark Datasets}
    In this paper, algorithms are validated using the Penn Machine Learning Benchmark (PMLB)~\cite{olson2017pmlb}, which is a curated collection of datasets from OpenML~\cite{vanschoren2014openml}. Our work focuses on real-world datasets. Thus, we exclude any datasets generated by the Friedman, Feynman, or Strogatz functions. After applying these exclusions, 58 datasets remain in the PMLB repository.

    \subsection{Parameter Settings}
    The parameter settings conform to the conventional settings of GP, as summarized in \cref{tab: parameter settings}. The analytical quotient (AQ)~\cite{ni2012use} replaces the division operator to prevent zero division errors. AQ is defined as $AQ=\frac{a}{\sqrt{1 + (b^2)}}$, where $a$ and $b$ are input variables. Also, the analytical log is used to replace the traditional log, which is defined as $Log=log(1+\sqrt{x^2})$ for input $x$. The sum of the probabilities for random mutation, random tree addition, and random tree deletion is set to 0.5 to explore the search space effectively~\cite{munoz2019evolving}. Elitism is only applied in the baseline algorithm when the environmental selection operator is not used, as the environmental selection operator in NSGA-II inherently supports elitism.

    \begin{table}[tb]
        \centering
        \caption{Parameter settings for SAM-GP.}
        \begin{tabular}{cc}%
            \toprule
            \textbf{Parameter}                 & \textbf{Value}                 \\%
            \midrule
            Maximal Population Size            & 200                            \\
            Number of Generations              & 100                            \\
            Crossover and Mutation Rates       & 0.9 and 0.1                    \\
            Tree Addition Rate                 & 0.5                            \\
            Tree Deletion Rate                 & 0.5                            \\
            Initial Tree Depth                 & 0-3                            \\
            Maximum Tree Depth                 & 10                             \\
            Initial Number of Trees            & 1                              \\
            Maximum Number of Trees            & 10                             \\
            Elitism (Number of Individuals)    & 1                              \\
            Standard Deviation of Perturbation & 0.3                            \\
            Iterations of Perturbation         & 10                             \\
            Functions                          & \makecell{+, -, *, AQ, Square, \\ Sqrt, Abs, Log, Max, Min, \\ Sin, Cos, Negative, Sigmoid}                          \\
            \bottomrule
        \end{tabular}%
        \label{tab: parameter settings}%
    \end{table}%

    \subsection{Performance Measure}
    To ensure the reliability of our conclusions, all experiments are conducted over 30 independent runs, each initialized with a different random seed. In each run, 100 samples are randomly selected from the dataset to be used as the training data, while the remaining samples are employed as the test data. This setup follows the pioneering work in studying the generalization of GP~\cite{ni2015training,nicolau2021choosing} and is designed to simulate scenarios with limited data availability. For datasets with fewer than 100 samples in either the training or test data, we split these datasets at a ratio of 50:50 to avoid having too little training and test data. For the experiments on label noise learning in \cref{sec: Performance on Label Noise}, the dataset is split into an 80:20 ratio for training and testing. Gaussian noise $\mathcal{N}(0,1)$ is added to the labels of the training data. To eliminate the influence of magnitude differences on the evaluation scores, we employ the $R^2$ metric, which is defined as $1-\frac{\sum_i (y_i-\hat{y}_i)^2}{\sum_i (y_i-\bar{y})^2}$, where $\hat{y}_i$ denotes the prediction of the model for data instance $i$, $y_i$ is the actual value for instance $i$, and $\bar{y}$ is the mean of the actual values. Finally, a statistical comparison of the algorithms is conducted using the Wilcoxon signed-rank test with a significance level of 0.05.

    In this study, we investigate both single-model and ensemble feature construction strategies. For single-model feature construction, the algorithm is referred to as \textit{SAM-GP}. As for ensemble feature construction, we set the ensemble size to 100 and refer to the algorithm as \textit{SAM-EGP}.

    \subsection{Baseline Methods}
    This paper evaluates the performance of SAM-GP against seven baseline GP algorithms with various overfitting control methods:
    \begin{itemize}
        \item Standard GP without Regularization (Standard GP): This is the standard GP algorithm that uses leave-one-out cross-validation (LOOCV) loss as the sole optimization objective.
        \item Parsimony Pressure (PP)~\cite{zhang1995balancing}: This approach uses the size of GP trees as a measure of complexity to control overfitting.
        \item Tikhonov Regularization (TK)~\cite{ni2014tikhonov}: This method employs zero-order Tikhonov, i.e., the L2 norm of the outputs of the model, as a measure of complexity. Zero-order Tikhonov is used because there is no significant difference between using different orders of Tikhonov regularization for GP, and it is computationally efficient~\cite{ni2014tikhonov}.
        \item Grand Complexity (GC)~\cite{ni2014tikhonov}: GC considers both model size and Tikhonov complexity, using their dominance rank to determine overall complexity.
        \item Rademacher Complexity (RC)~\cite{chen2020rademacher}: This is a data-dependent complexity measure that gauges the capacity of a machine learning model to fit a dataset. Formally, for a set of models $\mathcal{F}$, with corresponding loss functions denoted as $\mathcal{L}$, the Rademacher complexity is defined as follows:
        \begin{equation}
            \operatorname{R}_n(\mathcal{L})=\mathbb{E}\left[\sup _{l \in \mathcal{L}} \frac{1}{n} \sum_{i=1}^n \zeta_i l\left(x_i,y_i\right)\right],
        \end{equation}
        where $\zeta_i$ represents a randomly sampled Rademacher variable from $\{-1,1\}$. For regression, the Rademacher complexity could be unbounded due to the unbounded loss. Thus, for simplicity, we fit the machine learning model based on constructed features using target $- \zeta_i y_i$ to obtain an estimated Rademacher complexity. The loss function used in this paper is mean squared loss, and thus when $\zeta_i=1$, $l\left(x_i,y_i\right)$ will be optimized toward $(2 y_i)^2$, and when $\zeta_i=-1$, $l\left(x_i,y_i\right)$ will be optimized toward 0.
        \item Weighted MIC between Residuals and Variables (WCRV)~\cite{chen2020improving}: WCRV aims to minimize the high correlation between the chosen variables and residuals. Here, the maximum information coefficient (MIC) is used as the metric to measure the correlation between variables $x^k$ and residuals $R$. Specifically, the WCRV of an individual $\Phi$ is defined as:
        \begin{equation}
            \begin{split}
                \operatorname{WCRV}\left(\Phi\right)=&\sum_{\operatorname{MIC}_{x^k, Y}>=mv} \operatorname{MIC}_{x^k, Y}\times \operatorname{MIC}_{x^k, R} \\
                &+\sum_{\operatorname{MIC}_{x^k, Y}<mv}\left(1-\operatorname{MIC}_{x^k, Y}\right).
            \end{split}
        \end{equation}
        For the selected variable with high MIC with target variable $Y$ that surpasses the median value $mv$, i.e., $\sum_{\operatorname{MIC}_{x^k, Y}>=mv}$, the MIC between variables $x^k$ and residuals $R$ is summed. For the unimportant variable, the negative MIC between variables $x^k$ and target $Y$ is summed.
        \item Correlation between the Input and Output Distances (IODC)~\cite{vanneschi2021soft}: IODC first calculates pairwise Euclidean distances between inputs $D(x_i, x_j)$ and pairwise Euclidean distances between outputs $D(y_i, y_j)$, where $x_i$ and $x_j$ represent two arbitrary input points, and $y_i$ and $y_j$ represent two arbitrary output points. This results in two pairwise distance matrices for inputs and outputs, denoted as $I$ and $O$ respectively. Then, it measures the Pearson correlation between these two distance matrices using the formula:
        \begin{equation}
            \operatorname{IODC}(\Phi)=\frac{\operatorname{Cov}(\mathrm{I}, \mathrm{O})}{\sigma_{\mathrm{I}} \sigma_{\mathrm{O}}}
        \end{equation}
        where $Cov$ is the covariance matrix and $\sigma$ is the standard deviation. In this equation, a high correlation indicates better smoothness and may be less prone to overfitting.
    \end{itemize}
    All baseline methods follow the same framework as SAM-GP; the only difference lies in replacing the sharpness objective with the respective metrics of these methods. Bounded prediction is applied to all baseline methods to avoid extreme outlier predictions. For these baseline methods, existing work often selects the model with the best training error on the Pareto front as the final model~\cite{chen2020rademacher}. However, such a model can still be overly complex. Therefore, this paper uses the minimum Manhattan distance (MMD) method~\cite{chiu2016minimum} to determine the knee point as the final model. First, two objectives are normalized based on the best and worst objectives on the Pareto front, denoted as $(\mathcal{O}_1(\Phi), \mathcal{O}_2(\Phi))$, since these complexity measures do not have the same scale as mean square error. Then, MMD selects the point with the best sum of the two normalized objectives, i.e., $\mathcal{O}_1(\Phi)+\mathcal{O}_2(\Phi)$, as the final model.

    \section{Experimental Results}
    \label{Experimental Results}
    This section presents the experimental results of GP using SAM and six other overfitting control techniques, as well as the standard method for GP-based feature construction on 58 datasets, and compares the test performance, the training performance, the training time, and the model size.

    \subsection{Comparison on Test Performance}
    \begin{table*}[tb]
        \centering
        \caption{Statistical comparison of \textbf{test $R^2$ scores} when optimizing various model complexity measures.}
        \label{tab: Test R2}
        \resizebox{\textwidth}{!}{
            \begin{tabular}{cccccccc}%
                \toprule%
                & \textbf{PP}             & \textbf{RC}             & \textbf{GC}             & \textbf{IODC}            & \textbf{TK}              & \textbf{WCRV}            & \textbf{Standard GP}     \\%
                \midrule%
                \textbf{SAM}  & 32(+)/23($\sim$)/3({-}) & 51(+)/7($\sim$)/0({-})  & 37(+)/20($\sim$)/1({-})& 41(+)/16($\sim$)/1({-})& 50(+)/8($\sim$)/0({-})& 41(+)/13($\sim$)/4({-})& 36(+)/16($\sim$)/6({-})\\%
                \textbf{PP}   & ---                     & 36(+)/19($\sim$)/3({-}) & 15(+)/36($\sim$)/7({-}) & 23(+)/25($\sim$)/10({-})& 24(+)/33($\sim$)/1({-})& 20(+)/30($\sim$)/8({-})& 24(+)/26($\sim$)/8({-})\\%
                \textbf{RC}   & ---                     & ---                     & 2(+)/25($\sim$)/31({-}) & 4(+)/30($\sim$)/24({-})  & 4(+)/32($\sim$)/22({-})& 8(+)/24($\sim$)/26({-})& 15(+)/10($\sim$)/33({-})\\%
                \textbf{GC}   & ---                     & ---                     & ---                     & 22(+)/28($\sim$)/8({-})  & 25(+)/30($\sim$)/3({-})  & 17(+)/37($\sim$)/4({-})& 24(+)/21($\sim$)/13({-})\\%
                \textbf{IODC} & ---                     & ---                     & ---                     & ---                      & 21(+)/23($\sim$)/14({-}) & 19(+)/22($\sim$)/17({-}) & 23(+)/14($\sim$)/21({-}) \\%
                \textbf{TK}   & ---                     & ---                     & ---                     & ---                      & ---                      & 10(+)/27($\sim$)/21({-}) & 10(+)/27($\sim$)/21({-}) \\%
                \textbf{WCRV} & ---                     & ---                     & ---                     & ---                      & ---                      & ---                      & 18(+)/24($\sim$)/16({-}) \\%
                \bottomrule%
            \end{tabular}%
        }
    \end{table*}

    \subsubsection{General Analysis}
    \def\SAMLOOCVBetter{36}
    \def\SAMLOOCVWorse{6}
    \def\OtherLOOCVBetter{24}
    \def\SAMOtherBetter{32}
    The pairwise statistical comparison of test $R^2$ scores on 58 datasets is presented in \cref{tab: Test R2}, which compares six overfitting control methods with standard GP that uses standard GP as the sole objective function. In general, incorporating complexity measures can enhance generalization performance, although the degree of improvement varies, and some methods also degrade generalization on several datasets. Firstly, it is evident that SAM delivers outstanding generalization performance, significantly outperforming standard GP on \SAMLOOCVBetter{} datasets and underperforming on \SAMLOOCVWorse{} datasets out of 58, demonstrating the advantage of using SAM for better generalization. As for other methods, they outperform standard GP on at most \OtherLOOCVBetter{} datasets, indicating that their effectiveness is limited. Moreover, SAM outperforms other methods on more than \SAMOtherBetter{} datasets, indicating that SAM is a better method for overfitting control in general compared to other methods.

    \subsubsection{Sharpness vs. Model Size Control}
    \def\SAMPPBetter{32}
    \def\SAMPPWorse{3}
    One interesting phenomenon is that although PP is a poor overfitting control method for GP-based symbolic regression~\cite{vanneschi2010measuring}, it remains a strong competitor for controlling overfitting in GP-based evolutionary feature construction. There are two potential reasons. First, previous research in symbolic regression suggests that applying parsimony pressure can significantly reduce model accuracy due to the small model size~\cite{chen2020rademacher}. However, within the context of evolutionary feature construction, several small and weak trees can still produce accurate results if they complement each other~\cite{zhang2023semantic}. Second, choosing the ``knee point'' as the final model provides a good trade-off between overfitting and underfitting, contrasting with traditional approaches that select the model with the highest training accuracy as the final solution, which has the risk of overfitting due to excessive model complexity. These factors enable parsimony pressure to exhibit superior performance in overfitting control. Nonetheless, SAM still outperforms it on \SAMPPBetter{} datasets, while performing worse on only \SAMPPWorse{} datasets. Thus, we can conclude that constructing simple features can reduce overfitting to some extent, but considering the semantics of constructed features is a better way to control overfitting.

    \subsubsection{Sharpness vs. Extreme Value Control}
    \def\SAMTKBetter{50}
    \def\SAMGCBetter{37}
    If a model outputs an extremely large value for a single training sample, even if the mean squared error is small, the model may still be undesirable due to non-smoothness. This led to the development of Tikhonov regularization for overfitting control. However, the results show that SAM outperforms it on \SAMTKBetter{} datasets and is no worse on any dataset, indicating that Tikhonov regularization is not an effective method for overfitting control. In fact, the limited effectiveness of Tikhonov regularization in overfitting control for GP was previously acknowledged~\cite{ni2014tikhonov}. The deficiency of Tikhonov regularization led to the introduction of grand complexity, which combines Tikhonov regularization with parsimony pressure. However, the results demonstrate that although grand complexity outperforms Tikhonov regularization in overfitting control, SAM still outperforms it on \SAMGCBetter{} datasets. Based on the above analysis, we can conclude that controlling to avoid outputting extreme values is not enough to achieve good generalization performance.

    \subsubsection{Evolutionary Plots}
\begin{figure}[tb]
        \centering
        \includegraphics[width=\linewidth]{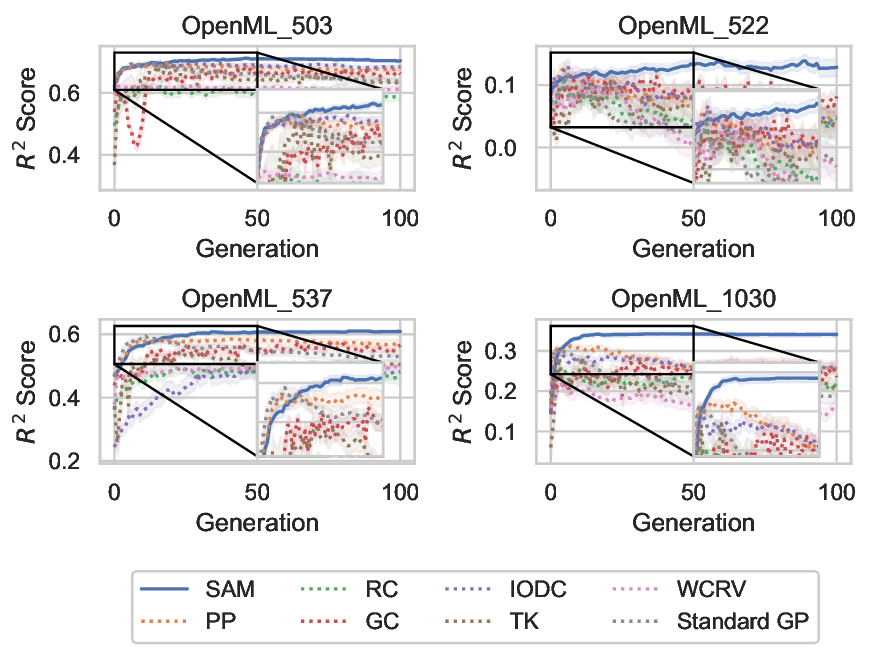}
        \caption{Evolutionary plots of \textbf{test $R^2$ scores} for different complexity control methods.}
        \label{fig: Test Plot}
    \end{figure}

    To gain a deeper understanding of the superior test performance of SAM, \cref{fig: Test Plot} shows the evolutionary plots of test $R^2$ scores on four representative datasets. The results show that using only LOOCV leads to overfitting of the training data in early generations, resulting in decreased test $R^2$ scores, as observed in ``OpenML\_503'' and `OpenML\_522.'' A similar situation can also be seen when using IODC and PP, although the degree of overfitting has been reduced. As for other methods, like GC and RC, they do not suffer from the issue of severe overfitting, but their test $R^2$ scores are stuck at a relatively low level. In contrast to these methods, SAM gradually improves $R^2$ scores over the evolutionary process and finally surpasses all baseline methods in the last generation. Combining these results, it is evident that SAM is a promising method that exhibits good performance in improving generalization on unseen data.

    \subsection{Comparison on Training Performance}
    \begin{table*}[tb]
        \centering
        \caption{Statistical comparison of \textbf{training $R^2$ scores} when optimizing various model complexity measures.}
        \label{tab: Training R2}
        \resizebox{\textwidth}{!}{
            \begin{tabular}{cccccccc}%
                \toprule%
                & \textbf{PP}              & \textbf{RC}            & \textbf{GC}             & \textbf{IODC}            & \textbf{TK}              & \textbf{WCRV}            & \textbf{Standard GP}   \\%
                \midrule%
                \textbf{SAM}  & 18(+)/19($\sim$)/21({-}) & 47(+)/7($\sim$)/4({-}) & 27(+)/24($\sim$)/7({-})& 25(+)/18($\sim$)/15({-})& 32(+)/13($\sim$)/13({-})& 27(+)/18($\sim$)/13({-})& 0(+)/4($\sim$)/54({-})\\%
                \textbf{PP}   & ---                      & 53(+)/5($\sim$)/0({-}) & 36(+)/22($\sim$)/0({-}) & 30(+)/22($\sim$)/6({-})& 33(+)/18($\sim$)/7({-})& 28(+)/22($\sim$)/8({-})& 0(+)/3($\sim$)/55({-})\\%
                \textbf{RC}   & ---                      & ---                    & 1(+)/12($\sim$)/45({-}) & 2(+)/13($\sim$)/43({-})  & 1(+)/4($\sim$)/53({-})& 2(+)/19($\sim$)/37({-})& 0(+)/0($\sim$)/58({-})\\%
                \textbf{GC}   & ---                      & ---                    & ---                     & 11(+)/34($\sim$)/13({-}) & 15(+)/26($\sim$)/17({-}) & 13(+)/32($\sim$)/13({-})& 0(+)/0($\sim$)/58({-})\\%
                \textbf{IODC} & ---                      & ---                    & ---                     & ---                      & 18(+)/17($\sim$)/23({-}) & 20(+)/17($\sim$)/21({-}) & 0(+)/1($\sim$)/57({-}) \\%
                \textbf{TK}   & ---                      & ---                    & ---                     & ---                      & ---                      & 25(+)/16($\sim$)/17({-}) & 0(+)/5($\sim$)/53({-}) \\%
                \textbf{WCRV} & ---                      & ---                    & ---                     & ---                      & ---                      & ---                      & 0(+)/4($\sim$)/54({-}) \\%
                \bottomrule%
            \end{tabular}%
        }
    \end{table*}

    \subsubsection{Overfitting of GP}
    \def\GPSAMTrainingBetter{54}
    To further confirm the presence of overfitting in GP, we present a pairwise comparison of training $R^2$ scores in \cref{tab: Training R2}. The experimental results indicate that optimizing using LOOCV as the sole objective leads to superior training performance, outperforming SAM in \GPSAMTrainingBetter{} out of the 58 datasets. However, when we analyze the training results in conjunction with the test results presented in \cref{tab: Test R2}, it becomes clear that relying solely on LOOCV results in overfitting the training data. In contrast, SAM, which exhibited significantly weaker training performance, achieved the best performance on the test data. The evolutionary plots for training $R^2$ depicted in \cref{fig: Training Plot} also support this observation. As illustrated, while optimizing LOOCV shows improved $R^2$ scores throughout the evolution process, \cref{fig: Test Plot} reveals that these improved training $R^2$ scores correspond to decreased test $R^2$ scores.

    \subsubsection{Alleviating Overfitting with Strong Regularization}
    When addressing severe overfitting, a common approach is to apply strong regularization. However, as indicated by \cref{fig: Training Plot}, the use of strong regularization significantly limits the training $R^2$ and adversely affects generalization performance. For instance, optimizing using RC introduces a strong regularization effect, resulting in notably lower training $R^2$ values compared to those achieved through optimizing LOOCV. However, this method does not exhibit good generalization performance because of underfitting. In other words, enhancing generalization requires careful selection of the correct implicit regularization bias. Excessive regularization, as demonstrated in these real-world problems, is counterproductive to generalization performance.

    \subsubsection{Alleviating Overfitting by SAM}
\begin{figure}[tb]
        \centering
        \includegraphics[width=\linewidth]{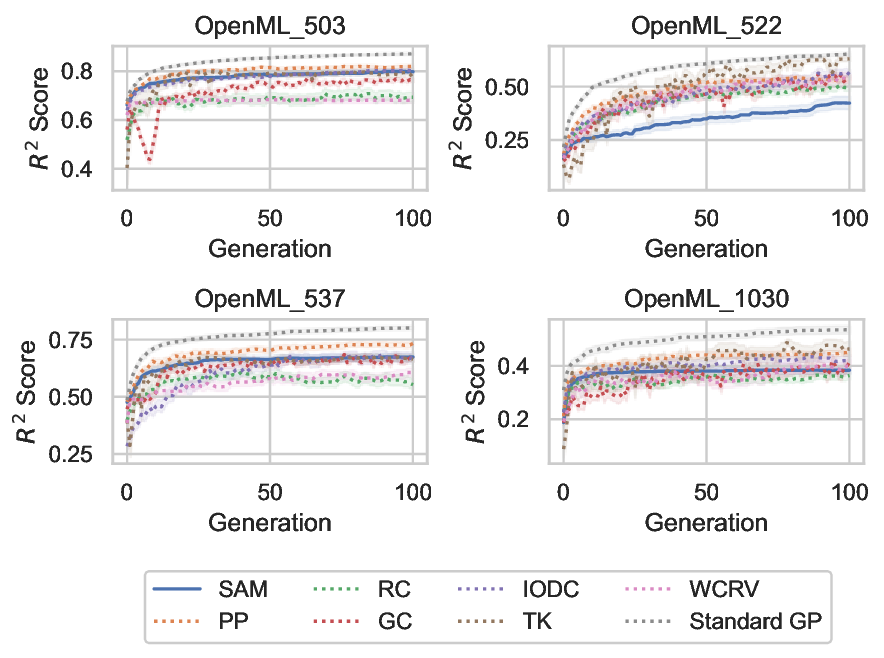}
        \caption{Evolutionary plots of \textbf{training $R^2$ scores} for various complexity control methods.}
        \label{fig: Training Plot}
    \end{figure}

\begin{figure}[tb]
        \centering
        \includegraphics[width=\linewidth]{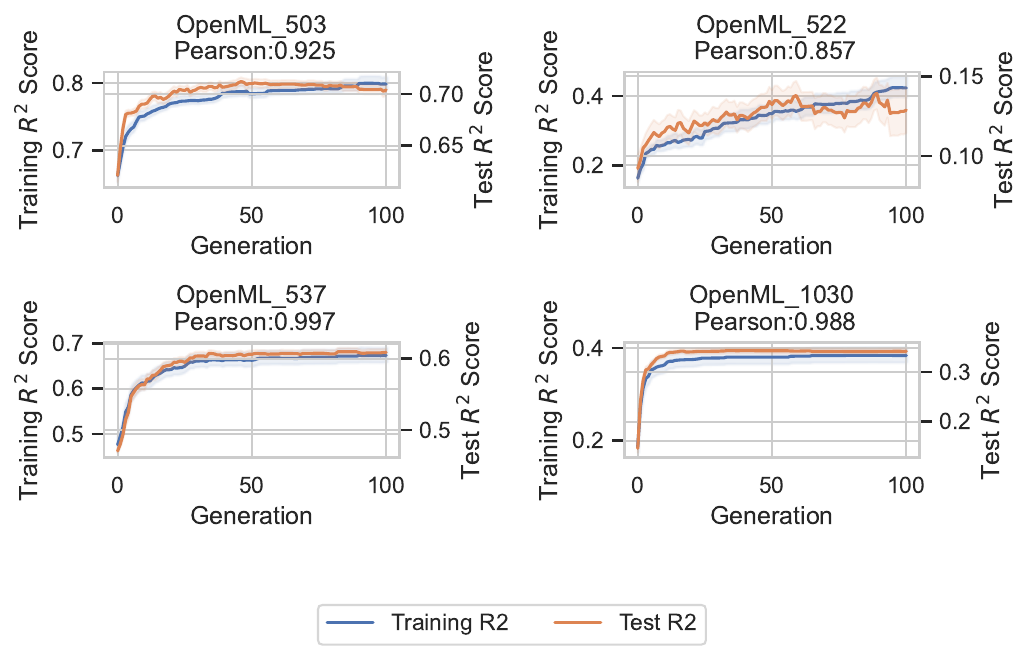}
        \caption{Evolutionary plots of \textbf{training and test $R^2$ scores} for SAM-GP.}
        \label{fig: Training Test Plot SAM}
    \end{figure}

\begin{figure}[tb]
        \centering
        \includegraphics[width=\linewidth]{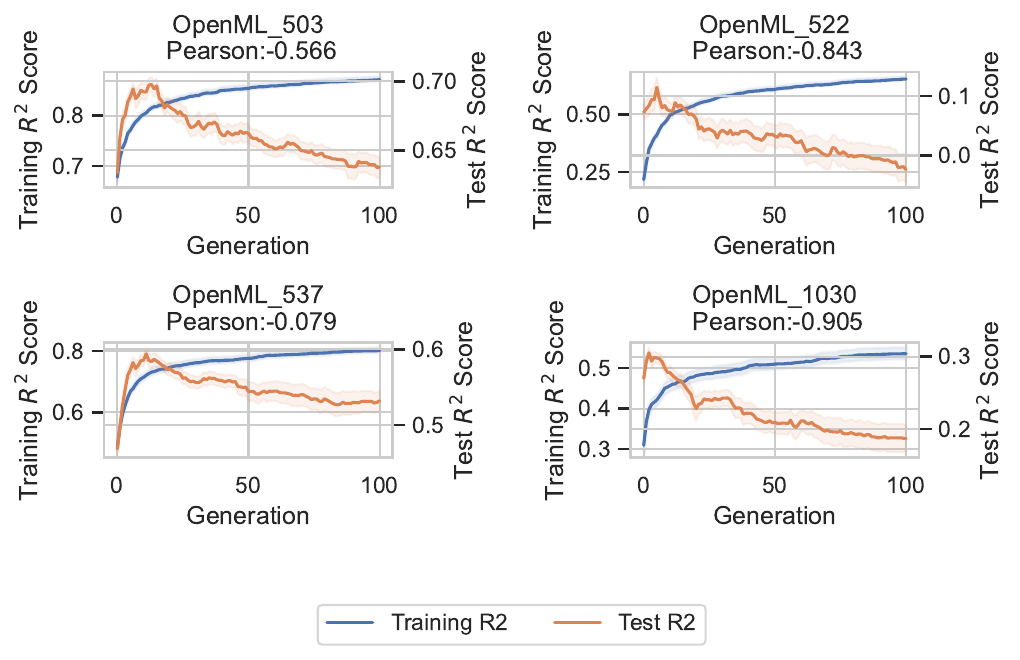}
        \caption{Evolutionary plots of \textbf{training and test $R^2$ scores} for GP without regularization.}
        \label{fig: Training Test Plot LOOCV}
    \end{figure}

    \cref{fig: Training Plot} and \cref{fig: Test Plot} provide a general view of the advantage of SAM over other complexity measures. To further demonstrate that SAM can effectively mitigate overfitting throughout the entire evolutionary process, \cref{fig: Training Test Plot SAM} illustrates the changes in training $R^2$ and test $R^2$ over the evolutionary process. There is a growing trend in both training and test $R^2$ scores as the generations increase. The Pearson correlations between training $R^2$ and test $R^2$ are also presented in \cref{fig: Training Test Plot SAM} and \cref{fig: Training Test Plot LOOCV}. The results in \cref{fig: Training Test Plot SAM} show a correlation range of 0.85 to 0.99 between training and test $R^2$ scores when using SAM, indicating a consistently strong and positive correlation throughout the optimization process. In comparison, the results in \cref{fig: Training Test Plot LOOCV} show a lower correlation between training and test $R^2$ scores, indicating the presence of overfitting. All these findings affirm that SAM is an effective strategy for optimizing GP in scenarios where overfitting is a concern.

    \subsection{Comparison of Training Time}
    \label{Comparison on Training Time}
\begin{figure}[tb]
        \centering
        \includegraphics[width=\linewidth]{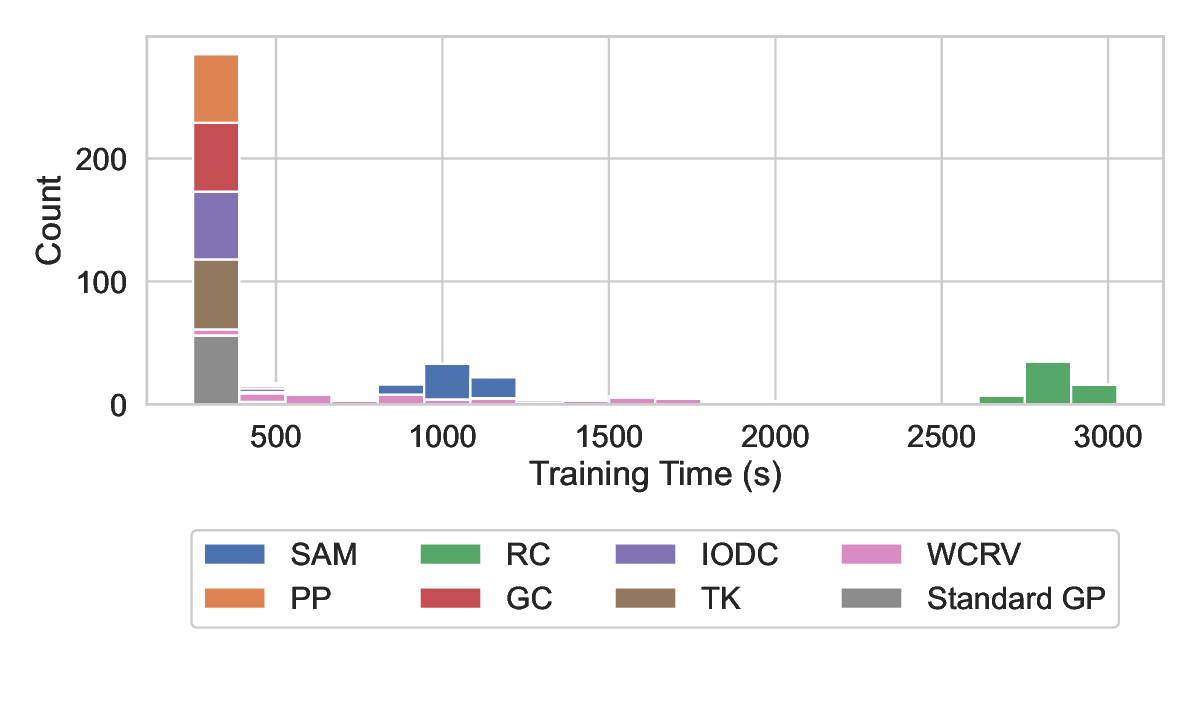}
        \caption{Distribution of \textbf{training time} across 58 datasets when optimizing different complexity measures.}
        \label{fig: Training Time}
    \end{figure}

    In addition to predictive performance, running time is also an important metric to consider when using a complexity measure for evolutionary feature construction in real-world applications. In \cref{fig: Training Time}, we present the distribution of running times for various measures. The results indicate that estimating sharpness is more time-consuming than complexity measures such as model sizes and grand complexity. Optimizing model sizes takes a similar amount of time compared to optimizing LOOCV, but optimizing SAM takes approximately three times longer for training compared to optimizing LOOCV alone or optimizing model sizes/grand complexity. This outcome was expected because grand complexity calculations are based solely on the structure of GP trees, whereas estimating sharpness requires calculating the change in the semantics of GP trees under perturbations, which is more time-consuming. Nonetheless, optimizing SAM is still more efficient than optimizing Rademacher complexity. More importantly, considering the significant improvement in generalization performance, the additional training time is deemed justifiable. The overfitting problem cannot be solved by simply increasing the number of generations in evolution, and it is still a fair comparison even though the training time is larger than standard GP.

    \subsection{Comparison of Tree Size}
\begin{figure}[!tb]
        \centering
        \includegraphics[width=\linewidth]{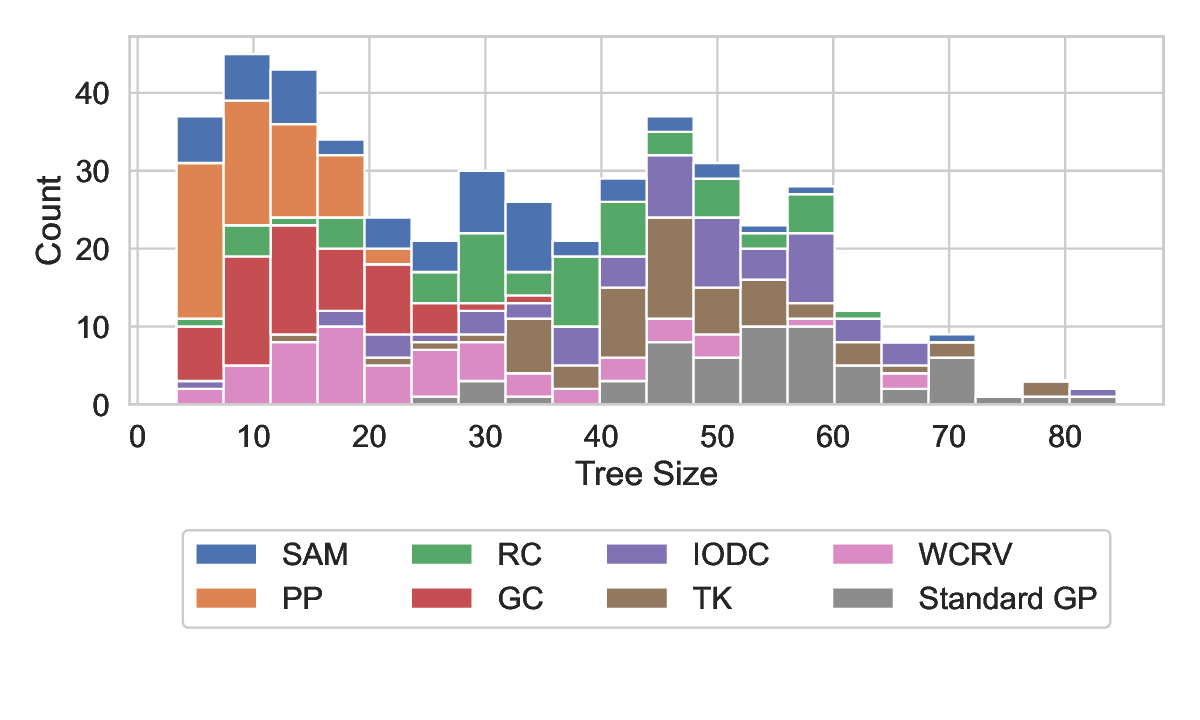}
        \caption{Distribution of \textbf{tree sizes} across 58 datasets when optimizing different model complexity measures.}
        \label{fig: Complexity}
    \end{figure}

    \def\SAMPPSizeWorse{41}
    \def\SAMPPSizeSimilar{10}
    \def\SAMPPSizeBetter{7}
    \def\SAMGPSizeWorse{0}
    \def\SAMGPSizeSimilar{6}
    \def\SAMGPSizeBetter{52}
    In this section, we present comparisons of the tree size of GP-constructed features. Although this aspect is not the primary focus of this paper, it is relevant to interpretability and generalization performance. The historical data related to each sharpness reduction layer is considered as one node. First, as depicted in \cref{fig: Complexity}, optimizing tree size or grand complexity results in the smallest trees; however, optimizing sharpness does not achieve comparable results because the tree size has not been explicitly optimized. Compared to PP, SAM results in significantly larger trees in \SAMPPSizeWorse{} datasets, yields similar results in \SAMPPSizeSimilar{} datasets, and achieves better results in only \SAMPPSizeBetter{} datasets. Although not comparable to PP, optimizing sharpness still moderately controls the model size compared to standard GP, even though tree size is not the direct target of optimization. Specifically, as shown in \cref{tab: Tree Size}, SAM results in significantly smaller trees in \SAMGPSizeBetter{} datasets compared to standard GP and yields similar results in \SAMGPSizeSimilar{} datasets. This demonstrates that optimizing sharpness can lead to more interpretable models compared to purely optimizing training loss. Nonetheless, it is essential to emphasize again that smaller models do not necessarily result in better generalization performance. For example, optimizing grand complexity and parsimony pressure results in small tree sizes but leads to poor generalization performance, as shown in \cref{tab: Test R2}. This further confirms existing findings in GP that a complexity measure relying solely on tree size is insufficient~\cite{vanneschi2010measuring}, and semantics need to be considered in the optimization to improve generalization performance.

    \begin{table*}[!tb]
        \centering
        \caption{Statistical comparison of \textbf{tree sizes} when optimizing different model complexity measures.}
        \label{tab: Tree Size}
        \resizebox{\textwidth}{!}{
            \begin{tabular}{cccccccc}%
                \toprule%
                & \textbf{PP}             & \textbf{RC}              & \textbf{GC}              & \textbf{IODC}           & \textbf{TK}             & \textbf{WCRV}            & \textbf{Standard GP}    \\%
                \midrule%
                \textbf{SAM}  & 7(+)/10($\sim$)/41({-}) & 24(+)/19($\sim$)/15({-}) & 10(+)/14($\sim$)/34({-})& 44(+)/6($\sim$)/8({-})& 46(+)/11($\sim$)/1({-})& 15(+)/17($\sim$)/26({-})& 52(+)/6($\sim$)/0({-})\\%
                \textbf{PP}   & ---                     & 54(+)/4($\sim$)/0({-})   & 32(+)/20($\sim$)/6({-})  & 57(+)/1($\sim$)/0({-})& 58(+)/0($\sim$)/0({-})& 44(+)/11($\sim$)/3({-})& 58(+)/0($\sim$)/0({-})\\%
                \textbf{RC}   & ---                     & ---                      & 3(+)/5($\sim$)/50({-})   & 32(+)/25($\sim$)/1({-}) & 35(+)/21($\sim$)/2({-})& 13(+)/8($\sim$)/37({-})& 45(+)/12($\sim$)/1({-})\\%
                \textbf{GC}   & ---                     & ---                      & ---                      & 56(+)/2($\sim$)/0({-})  & 58(+)/0($\sim$)/0({-})  & 38(+)/17($\sim$)/3({-})& 58(+)/0($\sim$)/0({-})\\%
                \textbf{IODC} & ---                     & ---                      & ---                      & ---                     & 8(+)/42($\sim$)/8({-})  & 5(+)/8($\sim$)/45({-})   & 22(+)/32($\sim$)/4({-}) \\%
                \textbf{TK}   & ---                     & ---                      & ---                      & ---                     & ---                     & 2(+)/9($\sim$)/47({-})   & 29(+)/26($\sim$)/3({-}) \\%
                \textbf{WCRV} & ---                     & ---                      & ---                      & ---                     & ---                     & ---                      & 54(+)/4($\sim$)/0({-})  \\%
                \bottomrule%
            \end{tabular}%
        }
    \end{table*}

    \subsection{Comparison with Other Machine Learning Algorithms}
\begin{figure*}[!tb]
        \centering
        \includegraphics[width=0.85\linewidth]{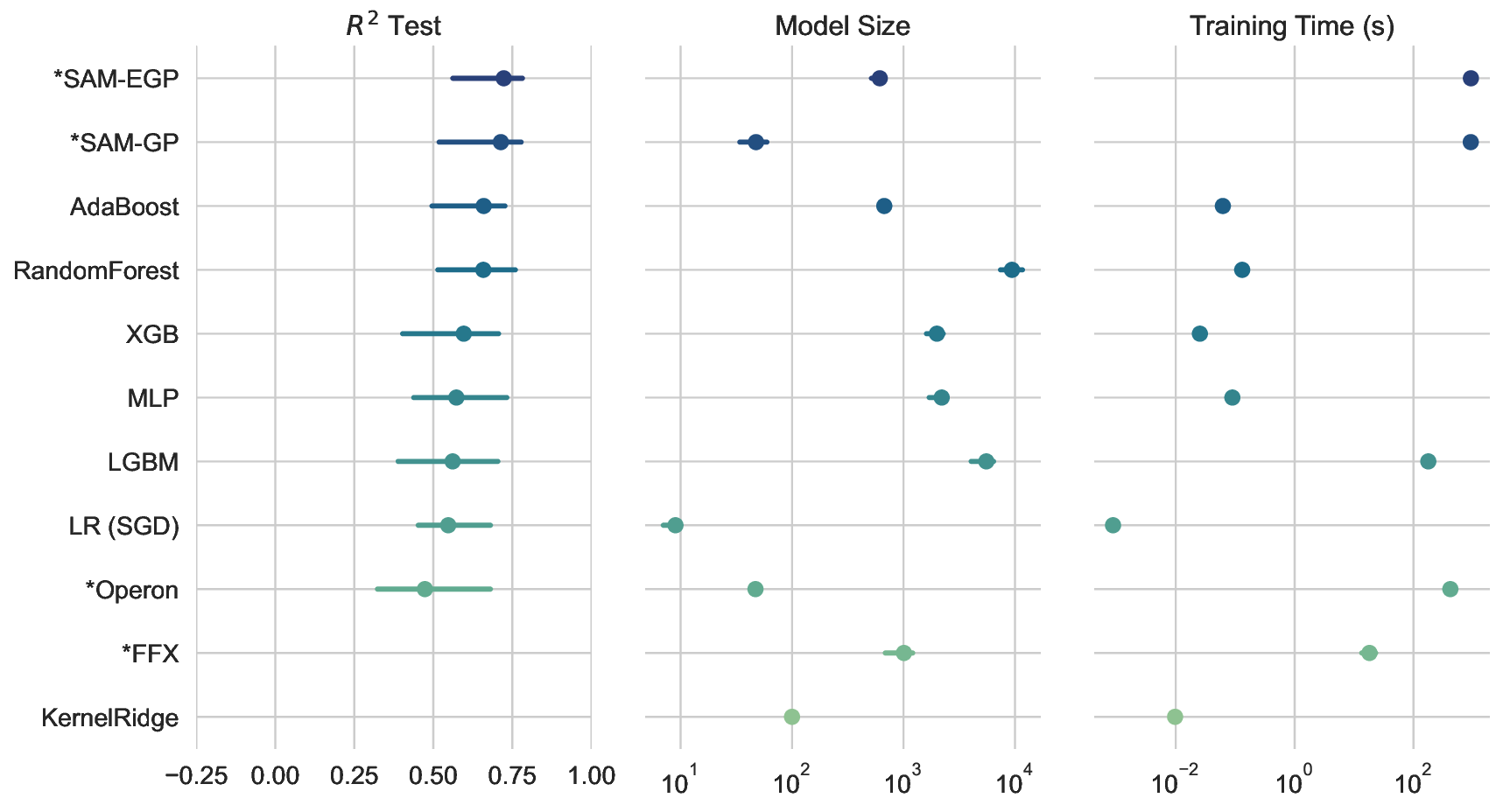}
        \caption{\textbf{Median $R^2$ scores, model sizes, and training time} of different algorithms on 58 regression problems.}
        \label{fig: SRBench}
    \end{figure*}

\begin{figure}[!tb]
        \centering
        \includegraphics[width=\linewidth]{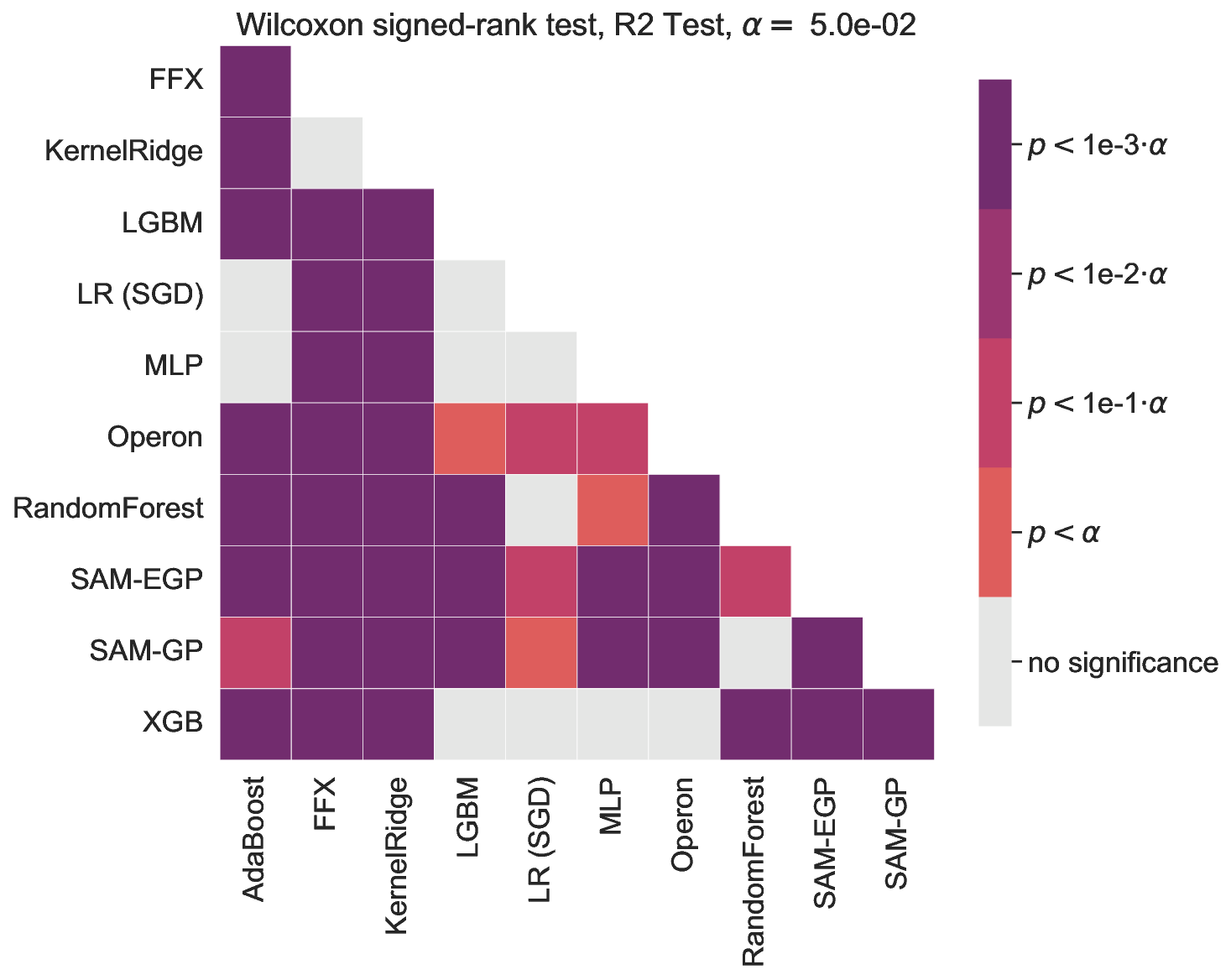}
        \caption{Pairwise statistical comparisons of \textbf{test $R^2$ scores} on 58 regression problems.}
        \label{fig: Pairwise Comparison}
    \end{figure}

    In addition to comparing different complexity measures, this paper investigates the performance of various machine learning algorithms, following the evaluation protocol of the state-of-the-art symbolic regression benchmark~\cite{cava2021contemporary}, but using the same training set split method as employed in this paper. The hyperparameters of all baseline algorithms are fine-tuned using cross-validation on the training data before model fitting. The evaluated algorithms encompass state-of-the-art machine learning methods, including XGBoost~\cite{DBLP:conf/kdd/ChenG16} and LightGBM~\cite{ke2017lightgbm}, as well as leading symbolic regression algorithms like Operon~\cite{burlacu2020operon}. Notably, due to the limited number of instances, hyperparameters for these machine learning algorithms are tuned using a grid search instead of successive halving grid search. The experimental results, including test $R^2$ scores, model sizes, and training times, are presented in \cref{fig: SRBench}. The results on test $R^2$ scores reveal that SAM-GP not only outperforms leading symbolic regression algorithms but also exceeds the performance of well-known machine learning algorithms in terms of generalization performance when data is limited. \cref{fig: Pairwise Comparison} shows the pairwise comparison of different algorithms using the Wilcoxon signed-rank test with Benjamini \& Hochberg correction. The results in \cref{fig: Pairwise Comparison} demonstrate that the advantage of SAM-GP over fine-tuned XGBoost is significant, indicating the superior generalization performance of SAM-GP. Furthermore, as shown in \cref{fig: SRBench} and \cref{fig: Pairwise Comparison}, the ensemble version of SAM-GP, i.e., SAM-EGP, significantly outperforms SAM-GP and fine-tuned XGBoost, as well as other machine learning algorithms in terms of test $R^2$ scores, further confirming the advantages of GP with SAM when dealing with limited data. Regarding model size, SAM-GP is one order of magnitude smaller than XGBoost, highlighting the superior interpretability of GP-based evolutionary feature construction methods. As for training time, SAM-GP takes longer than XGBoost but has a comparable training speed to Operon. While training time remains an issue that needs addressing in the future, given the superior predictive performance and interpretability of SAM-GP, it is a tradeoff that is acceptable in many machine-learning scenarios.

    \section{Further Analysis}
    \label{Further Analysis}

    \subsection{Analysis of Sharpness Reduction Layer}
\begin{figure}[!tb]
        \centering
        \includegraphics[width=\linewidth]{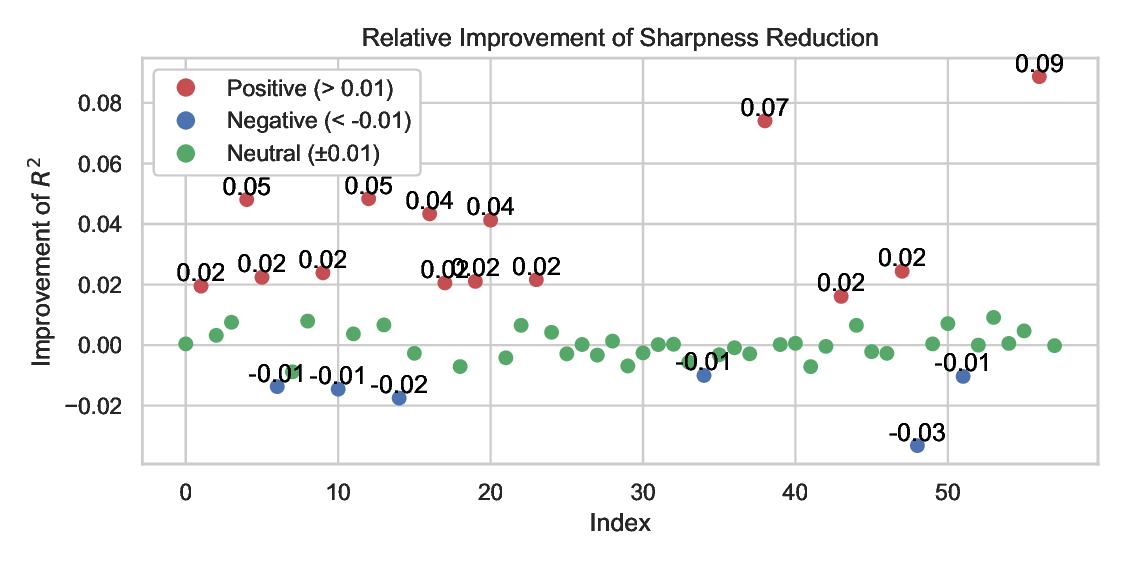}
        \caption{Improvements and decreases of more than 0.01 in $R^2$ scores brought by using the sharpness reduction layer, denoted by red and blue dots, respectively. The green dot represents neutral results with improvements and decreases of less than 0.01.}
        \label{fig: Sharpness Layer}
    \end{figure}

    \def\SharpnessLayerBetter{14}
    \def\SharpnessLayerWorse{6}
    \cref{fig: Sharpness Layer} shows the impact of the sharpness reduction layer on the test $R^2$ scores. For clarity, if improvements or decreases in $R^2$ are less than 0.01, they are simply denoted as neutral results, and the analysis is more focused on those datasets where $R^2$ could improve or decrease by more than 0.1. The results indicate that the sharpness reduction layer can improve $R^2$ by more than 0.01 on \SharpnessLayerBetter{} datasets compared to not using the sharpness reduction layer, and it leads to an $R^2$ decrease of more than 0.01 on only \SharpnessLayerWorse{} datasets. These results demonstrate that the sharpness reduction layer is an effective way to improve generalization performance.

    \subsection{Comparisons on Caching Technique}
\begin{figure}[!tb]
        \centering
        \includegraphics[width=\linewidth]{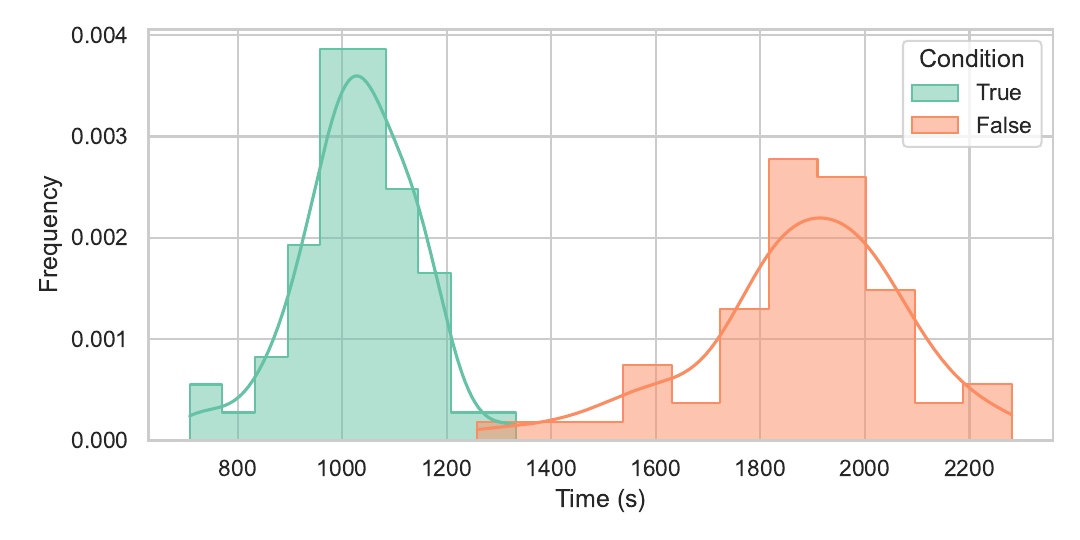}
        \caption{Distribution of training time (seconds) with and without the use of the caching technique.}
        \label{fig: Caching Time}
    \end{figure}

    \def\CacheTrue{1036}
    \def\CacheFalse{1893}
    \def\CacheImprovement{45}
    The experimental results investigating the effectiveness of the proposed caching technique for accelerating the learning process are presented in \cref{fig: Caching Time}. When using the caching technique, the median training time is \CacheTrue{}s. In contrast, without using the caching technique, the median training time is \CacheFalse{}s. In other words, after implementing the caching technique, the median training time is reduced by \CacheImprovement{}\%. These experimental results show that the proposed caching technique can significantly reduce the time required for sharpness estimation.

    \subsection{Analysis of Bounded Prediction}
\begin{figure}[tb]
        \centering
        \begin{subfigure}[b]{0.45\linewidth}
            \centering
            \includegraphics[width=\linewidth]{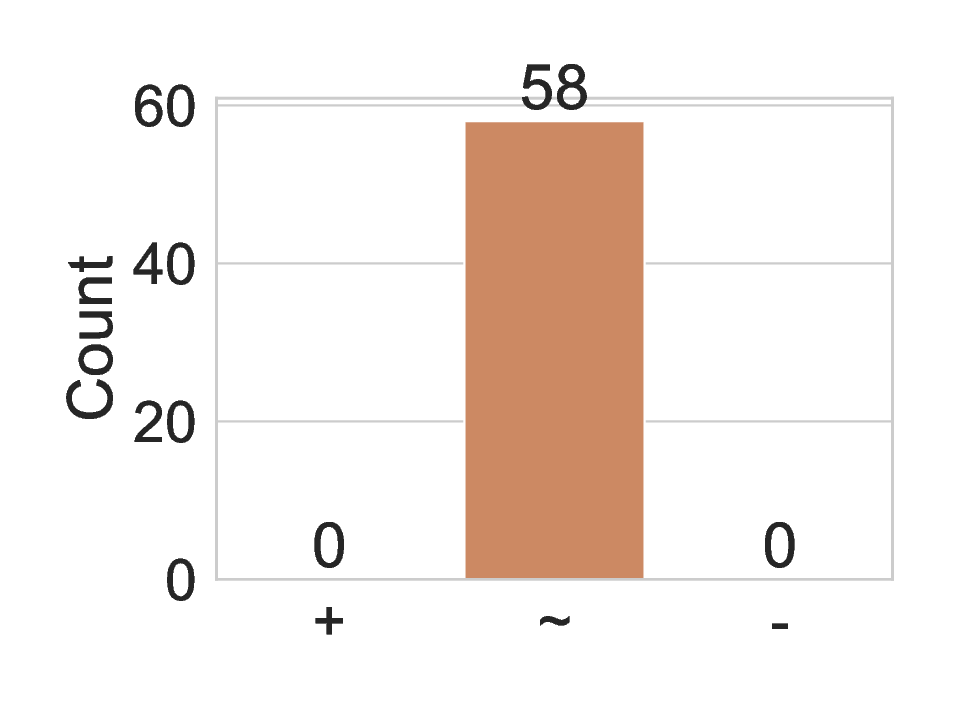}
            \caption{Training $R^2$}
            \label{fig: Bounded R2 Training}
        \end{subfigure}
        \begin{subfigure}[b]{0.45\linewidth}
            \centering
            \includegraphics[width=\linewidth]{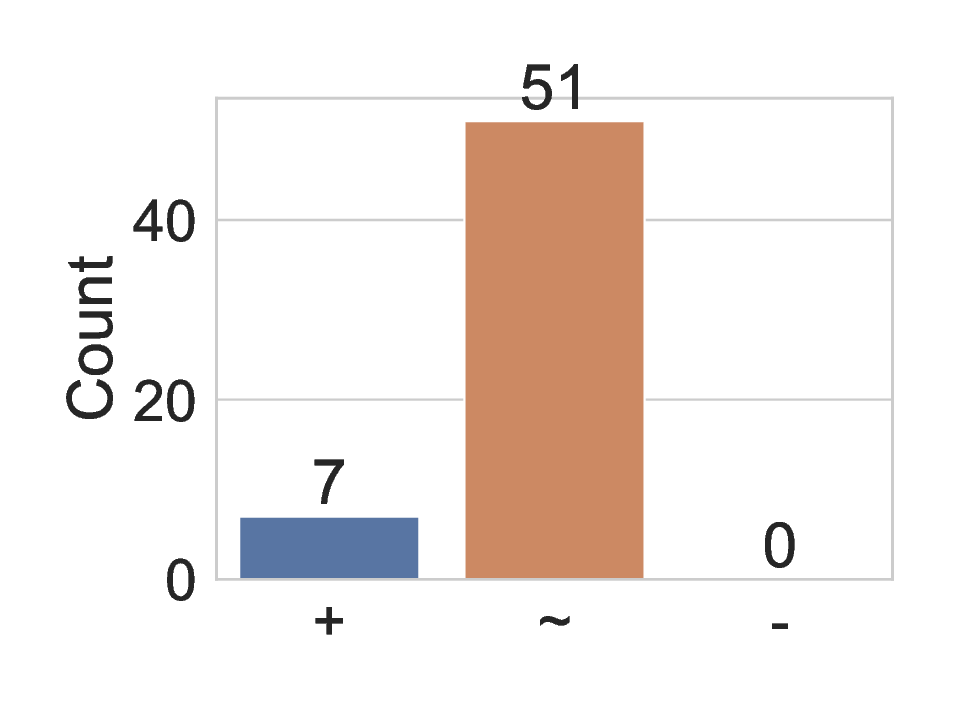}
            \caption{Test $R^2$}
            \label{fig: Bounded R2 Test}
        \end{subfigure}
        \caption{Statistical comparison of \textbf{training and test $R^2$ scores} before and after using bounded predictions.}
    \end{figure}

\begin{figure}[tb]
        \centering
        \includegraphics[width=\linewidth]{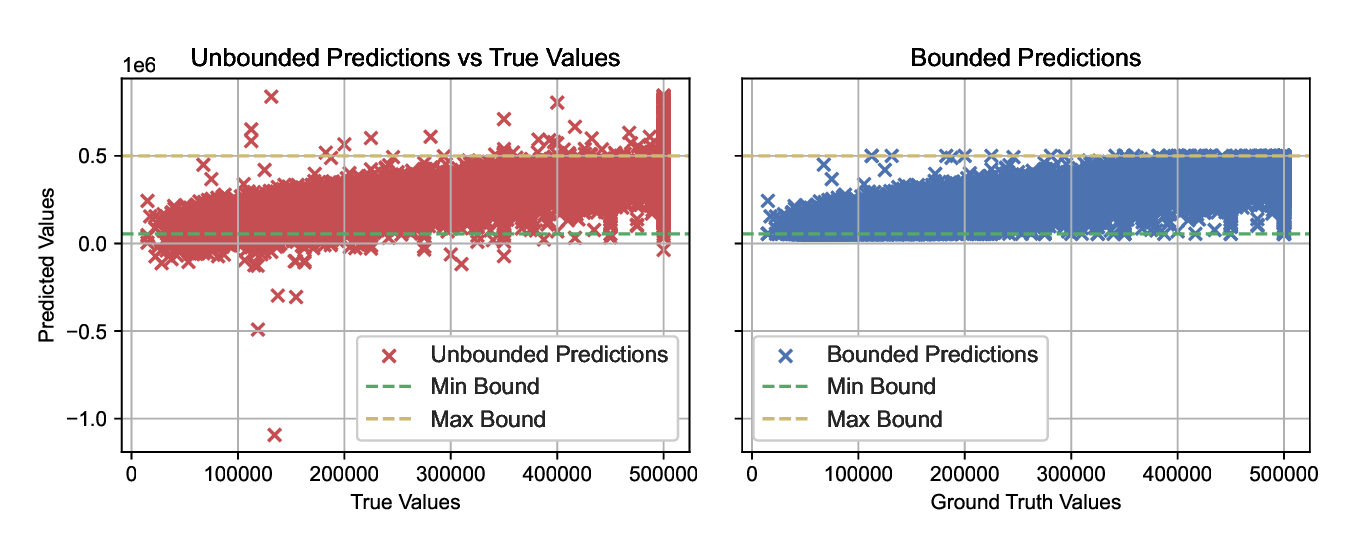}
        \caption{Effect of bounded predictions on ``OpenML 537'' test data.}
        \label{fig: Bounded Predictions}
    \end{figure}

    \def\BoundedPredictionBetter{7}
    In this paper, we introduce a bounded prediction mechanism to further reduce sharpness and improve generalization performance. As depicted in \cref{fig: Bounded R2 Test}, employing this mechanism along with SAM significantly improves generalization performance on \BoundedPredictionBetter{} datasets without reducing performance on any dataset. This suggests the effectiveness of bounded predictions in GP-based evolutionary feature construction. \cref{fig: Bounded R2 Training} presents a statistical test comparison on training performance, revealing that the proposed bounded prediction strategy does not significantly impact training performance, which aligns with the intuition that bounded prediction mainly restricts unreasonable extrapolation on unseen data. To further understand the impact of bounded predictions,  a case study on ``OpenML 537'' is depicted in \cref{fig: Bounded Predictions}. This figure shows that the model underestimates and overestimates predictions near the left and right boundaries of the prediction interval, respectively. The clipping strategy makes the model provide conservative predictions, thus achieving better prediction performance. The effectiveness of this clipping strategy in regression echoes a recent success in controlling overfitting by clipping logits in classification models~\cite{wei2023mitigating}. In the future, this issue could be more effectively addressed by finding a method to measure the KL divergence of GP models to control the KL divergence term in the PAC-Bayesian bound, i.e., $\frac{\|\mathbf{w}\|_2^2}{2 \sigma^2}$, to guide the model in avoiding predictions that deviate excessively from prior knowledge and further enhance generalization performance.

    \subsection{Performance in the Presence of Label Noise}
    \label{sec: Performance on Label Noise}
\begin{figure}[!tb]
        \centering
        \includegraphics[width=\linewidth]{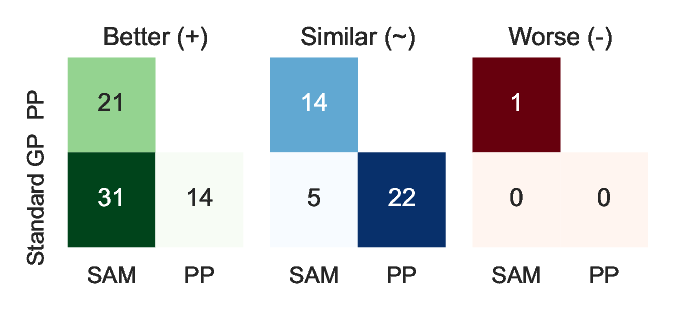}
        \caption{Statistical comparison of \textbf{test $R^2$ scores} on 36 datasets with label noise.}
        \label{fig: Label Noise Learning}
    \end{figure}

    \def\SAMGPNoiseBetter{31}
    \def\SAMPPNoiseBetter{21}
    In addition to limited samples, another scenario in evolutionary feature construction that poses a risk of overfitting is label noise. In the domain of deep learning, sharpness-aware minimization has demonstrated effectiveness in addressing label noise~\cite{foret2020sharpness}. In this section, we investigate the effectiveness of the proposed SAM-GP in label noise scenarios. Due to limited computational resources, the experiments are conducted on 36 datasets with a maximum of 2000 instances, and only standard GP and the top-performing method, i.e., PP, are compared. The experimental results are shown in \cref{fig: Label Noise Learning}. The results demonstrate that SAM improves generalization performance compared to standard GP on \SAMGPNoiseBetter{} datasets and outperforms PP on \SAMPPNoiseBetter{} datasets, without showing worse performance on any dataset. These results indicate that using sharpness-aware minimization in evolutionary feature construction is beneficial when labels contain noise, consistent with the findings in the deep learning domain~\cite{foret2020sharpness}.

    \subsection{Visualization of Constructed Features}
\begin{figure}[!tb]
        \centering
        \begin{subfigure}[b]{\linewidth}
            \centering
            \includegraphics[width=\linewidth]{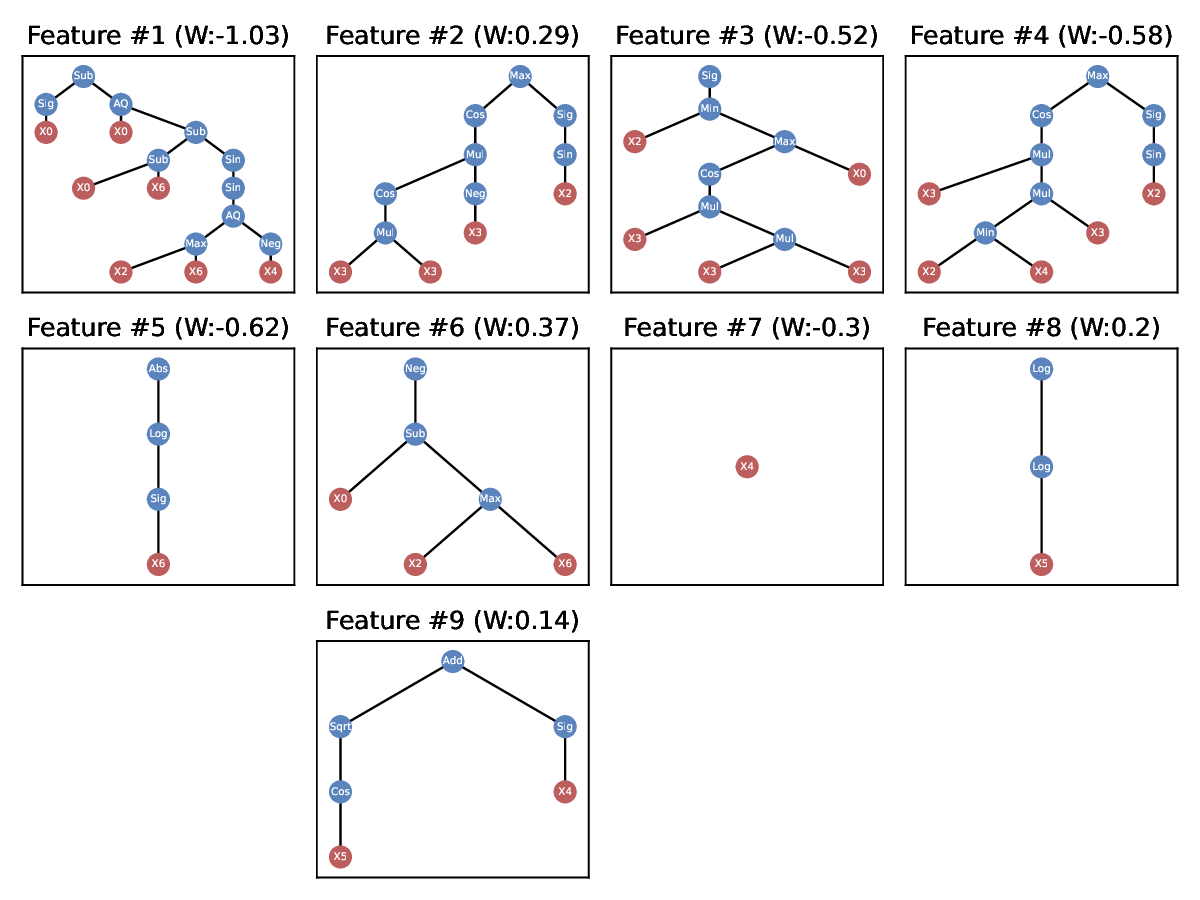}
            \caption{No Regularization}
            \label{fig: Illustrative Example PP}
        \end{subfigure}
        \begin{subfigure}[b]{\linewidth}
            \centering
            \includegraphics[width=\linewidth]{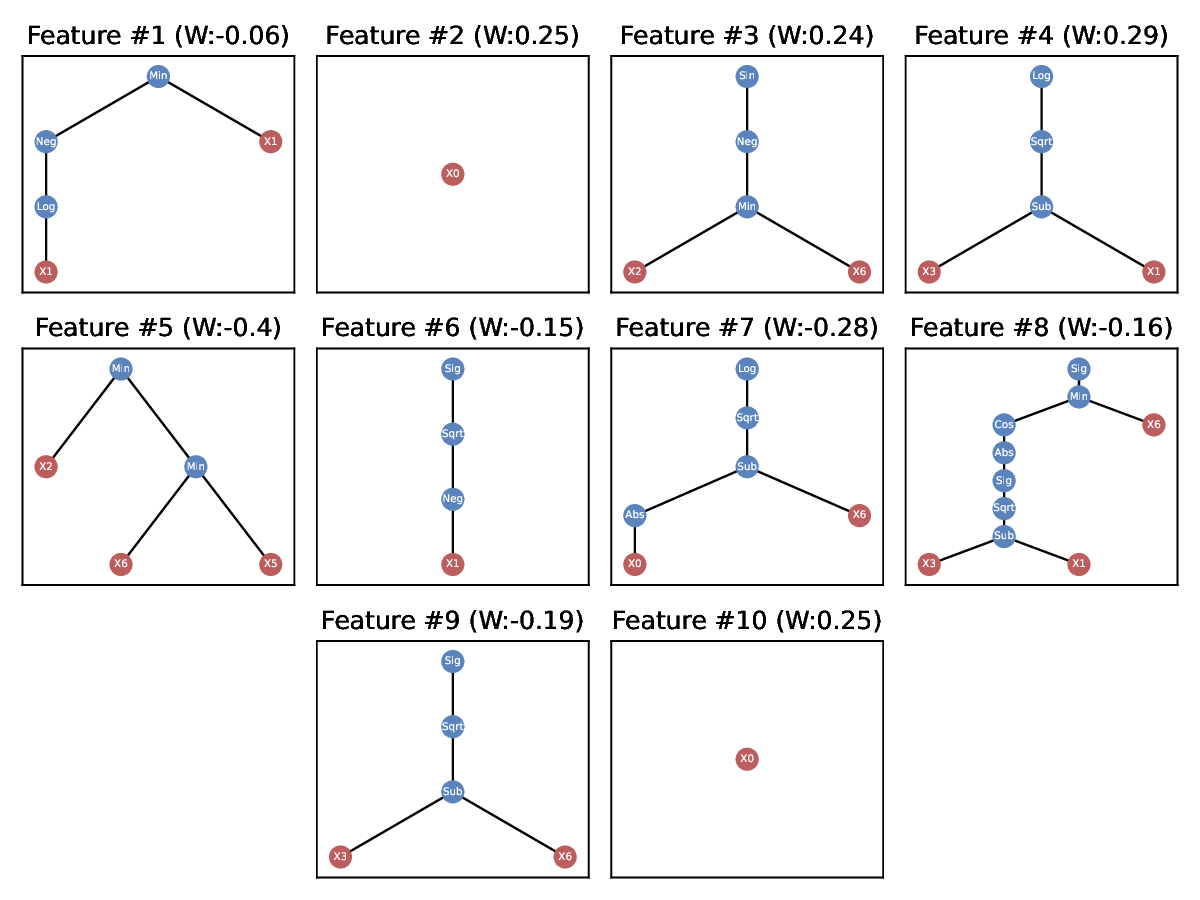}
            \caption{Sharpness-Aware Minimization}
            \label{fig: Illustrative Example SAM}
        \end{subfigure}
        \caption{Examples of constructed features based on (a). no regularization (b). sharpness-aware minimization.}
    \end{figure}

    \def\STDTraining{0.22}
    \def\STDTest{0.0}
    \def\SAMTraining{0.16}
    \def\SAMTest{0.18}
    In this paper, we have demonstrated the superior performance of SAM-GP. In this section, we investigate the features constructed by SAM-GP and standard GP to gain a deeper understanding of the differences between controlling overfitting by optimizing sharpness or not. An example of features optimized by standard GP and SAM-GP on ``OpenML 522'' is illustrated in \cref{fig: Illustrative Example PP}. Experimental results reveal that the features constructed using standard GP are complex, including nested functions like $sin(sin())$. In contrast, the model obtained through SAM is less functionally complex. In terms of training performance, standard GP can lead to a training performance of \STDTraining{}, while the training $R^2$ of the model obtained through SAM is \SAMTraining{}. However, regarding test performance, SAM achieves a test $R^2$ score of \SAMTest{}, whereas the model obtained by standard GP only has a test $R^2$ score of \STDTest{}, indicating overfitting in the model learned through standard GP and good generalization performance of the model learned by SAM.

    \section{Conclusions}
    \label{Conclusion}
    To enhance the generalization ability of evolutionary feature construction, this paper proposes a sharpness-aware minimization technique for symbolic models to control the sharpness of GP-constructed features and generate features that generalize well to unseen data. The experimental results on 58 datasets demonstrate the effectiveness of the proposed method in improving the generalization performance of constructed features compared to solely optimizing cross-validation loss or combining well-known complexity measures in statistical machine learning. This improvement surpasses that achieved by optimizing six other complexity measures. When comparing the optimization of SAM with other metrics, we find that improving generalization is more about achieving a flat minimum in the semantic space rather than optimizing model size. Additionally, the experimental results indicate that SAM-GP significantly outperforms fine-tuned state-of-the-art machine learning algorithms on limited samples, further demonstrating the effectiveness of GP-based learning algorithms.

    In this work, SAM is not the best option for all datasets. Therefore, in the future, it is worthwhile to determine the most suitable situations for using SAM based on the characteristics of each dataset, such as employing the instance space analysis technique~\cite{smith2020revisiting}. Another potential future direction is to validate the proposed technique with other AutoML approaches. The proposed method is not limited to evolutionary-based methods but is also applicable to Monte Carlo tree search-based AutoML methods~\cite{wever2021automl} and Bayesian optimization-based methods~\cite{zimmer2021auto}. Finally, it is also worth exploring the applicability of the proposed method to various symbolic regression algorithms, including reinforcement learning-based~\cite{mundhenk2021symbolic} and Transformer-based~\cite{kamienny2022end} symbolic regression techniques.


%

    \appendices

    \ifCLASSOPTIONcaptionsoff
    \newpage
    \fi



%
    \bibliographystyle{IEEEtranN}
    \bibliography{mybibliography}

    \begin{IEEEbiography}[{\includegraphics[width=1in,height=1.25in,clip,keepaspectratio]{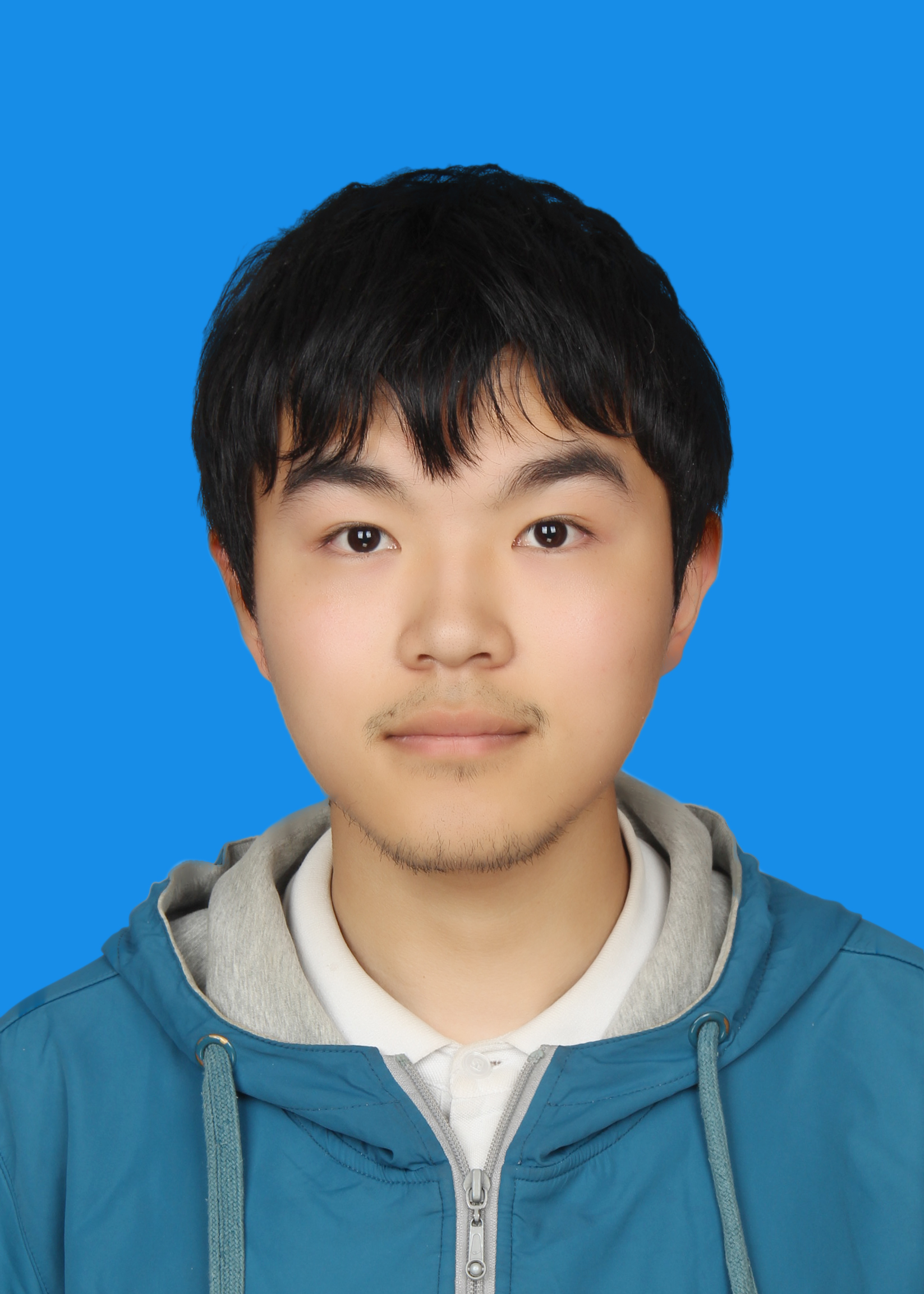}}]{Hengzhe Zhang}(M’23)
        is a PhD student in computer science at Victoria University of Wellington, New Zealand. He received the B.Sc. degree in software engineering from Xiangtan University, Hunan, China, in 2019, and the M.Sc. degree in computer science from East China Normal University, Shanghai, China, in 2022.

        His current research interests include symbolic regression, genetic programming, evolution computation, and statistical machine learning.
    \end{IEEEbiography}
    \vskip -2.5\baselineskip plus -1fil

    \begin{IEEEbiography}[{\includegraphics[width=1in,height=1.25in,clip,keepaspectratio]{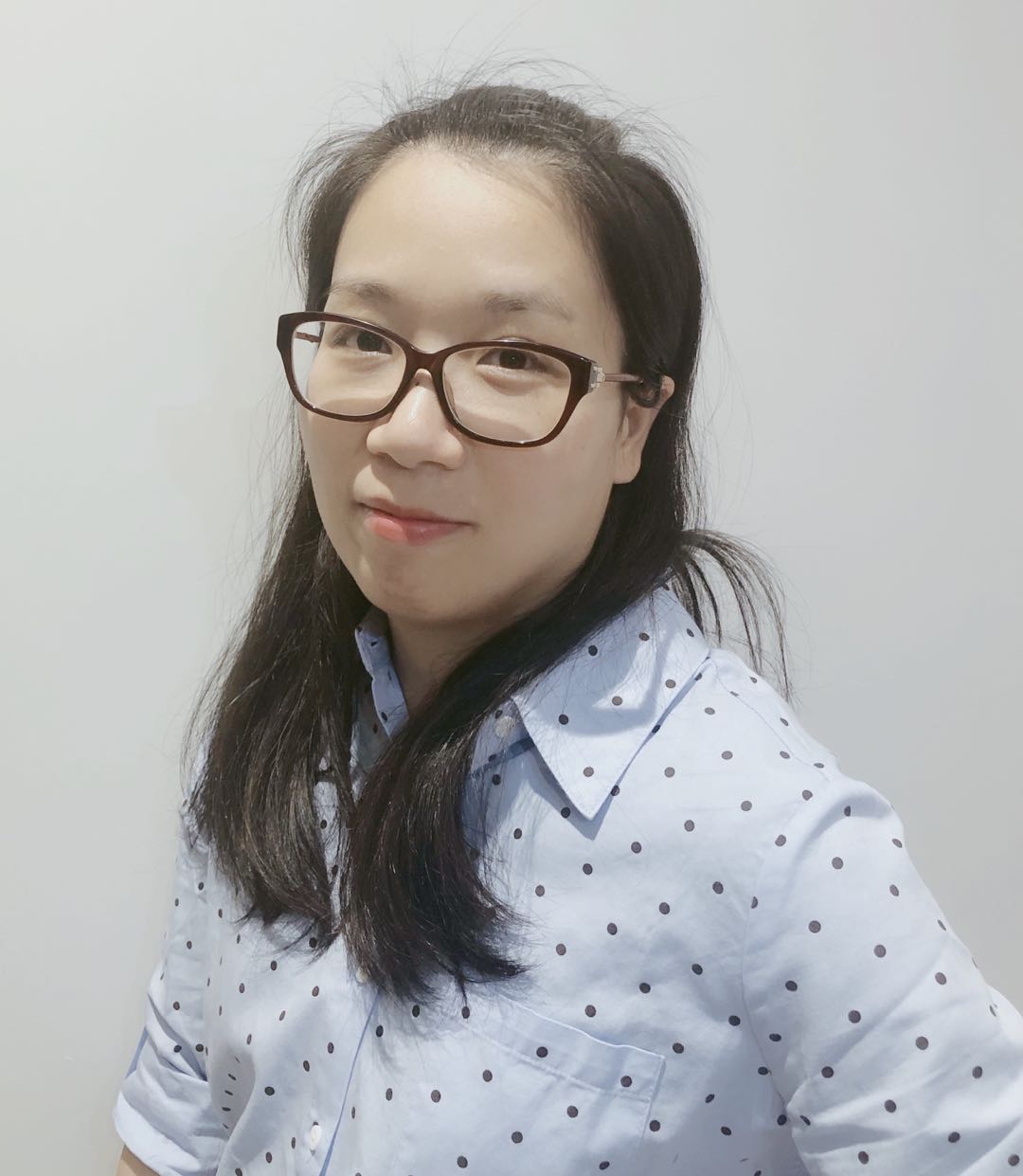}}]{Qi Chen} (M'14)
        received the B.E. degree in Automation from the University of South China, Hunan, China in 2005 and the M.E. degree in Software Engineering from Beijing Institute of Technology, Beijing, China in 2007, and the PhD degree in computer science in 2018 at Victoria University of Wellington, New Zealand. Currently, she is a Senior Lecturer in Artificial Intelligence in School of Engineering and Computer Science at VUW. Qi's research interests including machine learning, evolutionary computation, feature selection, feature construction, transfer learning, domain adaptation and statistical learning theory. She serves as a reviewer of international conferences, including AAAI and IJCAI, and international journals, including IEEE Transactions on Evolutionary Computation and IEEE Transactions on Cybernetics.
    \end{IEEEbiography}
    \vskip -2.5\baselineskip plus -1fil

    \begin{IEEEbiography}[{\includegraphics[width=1in,height=1.25in,clip,keepaspectratio]{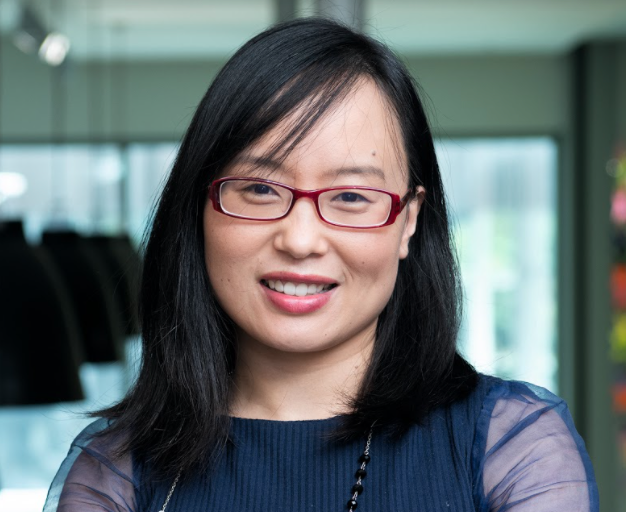}}]{Bing Xue} (M’10-SM’21)
        received the Ph.D. degree in computer science from the Victoria University of Wellington, Wellington, New Zealand, in 2014. She is currently a Professor of Artificial Intelligence, Deputy Director of Centre for Data Science and Artificial Intelligence, and Deputy Head of School in the School of Engineering and Computer Science at Victoria University of Wellington. She has over 300 papers published in fully refereed international journals and conferences and her research focuses mainly on evolutionary computation and machine learning. Dr Xue is currently the Chair of IEEE CIS Evolutionary Computation Technical Committee and Editor of IEEE CIS Newsletter. She has also served as associate editor of several international journals, such as IEEE CIM, IEEE TEVC and ACM TELO. She is also a Fellow of Engineering New Zealand.

    \end{IEEEbiography}
    \vskip -2.5\baselineskip plus -1fil

    \begin{IEEEbiography}[{\includegraphics[width=1in,height=1.25in,clip,keepaspectratio]{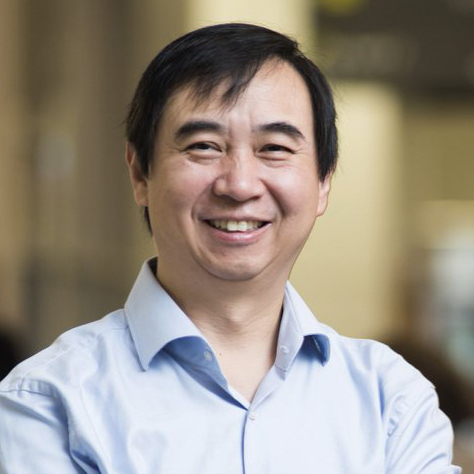}}]{Mengjie Zhang} (M’04-SM’10-F’19)
        received the Ph.D. degree in computer science from RMIT University, Melbourne, VIC, Australia, in 2000. He is currently a Professor of Computer Science, the Director of Centre for Data Science and Artificial Intelligence, Victoria University of Wellington, New Zealand. His current research interests include genetic programming, image analysis, feature selection and reduction, job-shop scheduling, and evolutionary deep learning and transfer learning. He has published over 800 research papers in refereed international journals and conferences. He is a Fellow of the Royal Society of New Zealand, a Fellow of Engineering New Zealand, a Fellow of IEEE, and an IEEE Distinguished Lecturer.
    \end{IEEEbiography}
    \vskip -2.5\baselineskip plus -1fil

    \begin{IEEEbiography}[{\includegraphics[width=1in,height=1.25in,clip,keepaspectratio]{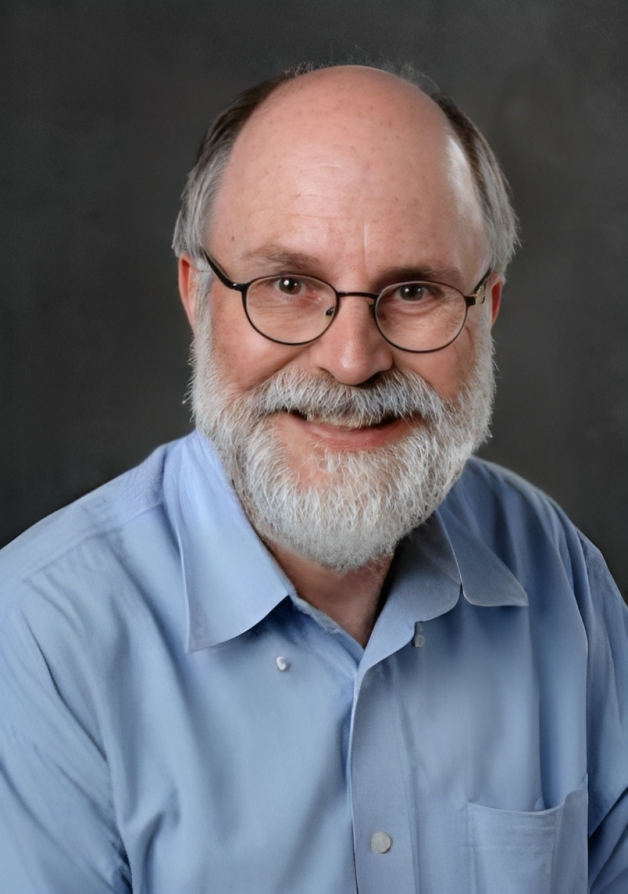}}]{Wolfgang Banzhaf}(M’20)
        received the Dr.rer.nat (Ph.D.) degree from the Department of Physics, Technische Hochschule Karlsruhe (currently, Karlsruhe Institute of Technology), Karlsruhe, Germany, in 1985. He was the University Research Professor with the Department of Computer Science, Memorial University of Newfoundland, St. John’s, NL, Canada, where he served as the Head of Department from 2003 to 2009 and from 2012 to 2016. He is the John R. Koza Chair of Genetic Programming with the Department of Computer Science and Engineering and a member of the BEACON Center for the Study of Evolution in Action, Michigan State University, East Lansing, MI, USA. Studies of selforganization and the field of Artificial Life are also of very much interest to him. He has become more involved with network research as it applies to natural and man-made systems. His research interests are in the field of bioinspired computing, notably evolutionary computation, and complex adaptive systems.
    \end{IEEEbiography}

%




    \clearpage

    \section{Empirically Verify Sharpness Estimation}
\begin{figure}[!htb]
        \centering
        \includegraphics[width=\linewidth]{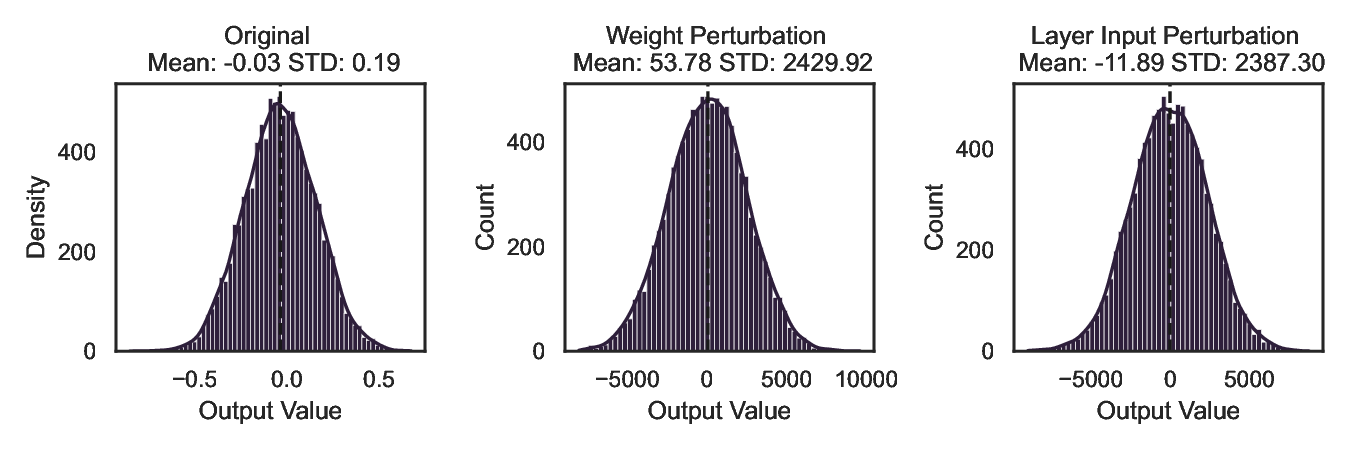}
        \caption{Distribution of outputs for different perturbation strategies.}
        \label{fig: Empirical Verification}
    \end{figure}

    The theorem in \cref{Theoretical Analysis} shows the equivalence between adding noise to input layers and adding noise to weights. To verify this, we tested it on a neural network with one hidden layer of 2000 neurons. The network is randomly initialized, where weight $w$ is drawn from a normal distribution $\mathcal{N}(0,1)$, and noise $\epsilon$ and $\epsilon'$ are drawn from normal distributions $\mathcal{N}(0,w_k^2)$ and $\mathcal{N}(0,x_i^2)$, respectively, and added to the weights $w_k$ and inputs of each layer $x_i$, i.e., $w_k'=w_k+\epsilon$ or $x_i'=x_i+\epsilon'$. To ensure reliable results, experiments are executed 100 times, each round with 100 randomly generated inputs with 2000 dimensions. The results are presented in \cref{fig: Empirical Verification}. They show that the distribution of outputs for weight perturbation and input perturbation is similar, thus empirically verifying the theorem in \cref{Theoretical Analysis}.

    \section{Hyperparameter Analysis}
    For SAM-GP, two hyperparameters need to be considered: the first one is the number of perturbations to estimate the sharpness, and the second one is the magnitude of perturbation added to the semantics of subtrees. This section performs a hyperparameter analysis to study the impact of these hyperparameters on the results.

    \subsection{Analysis of the Number of Perturbation Iterations}
\begin{figure}[tb]
        \centering
        \includegraphics[width=0.6\linewidth]{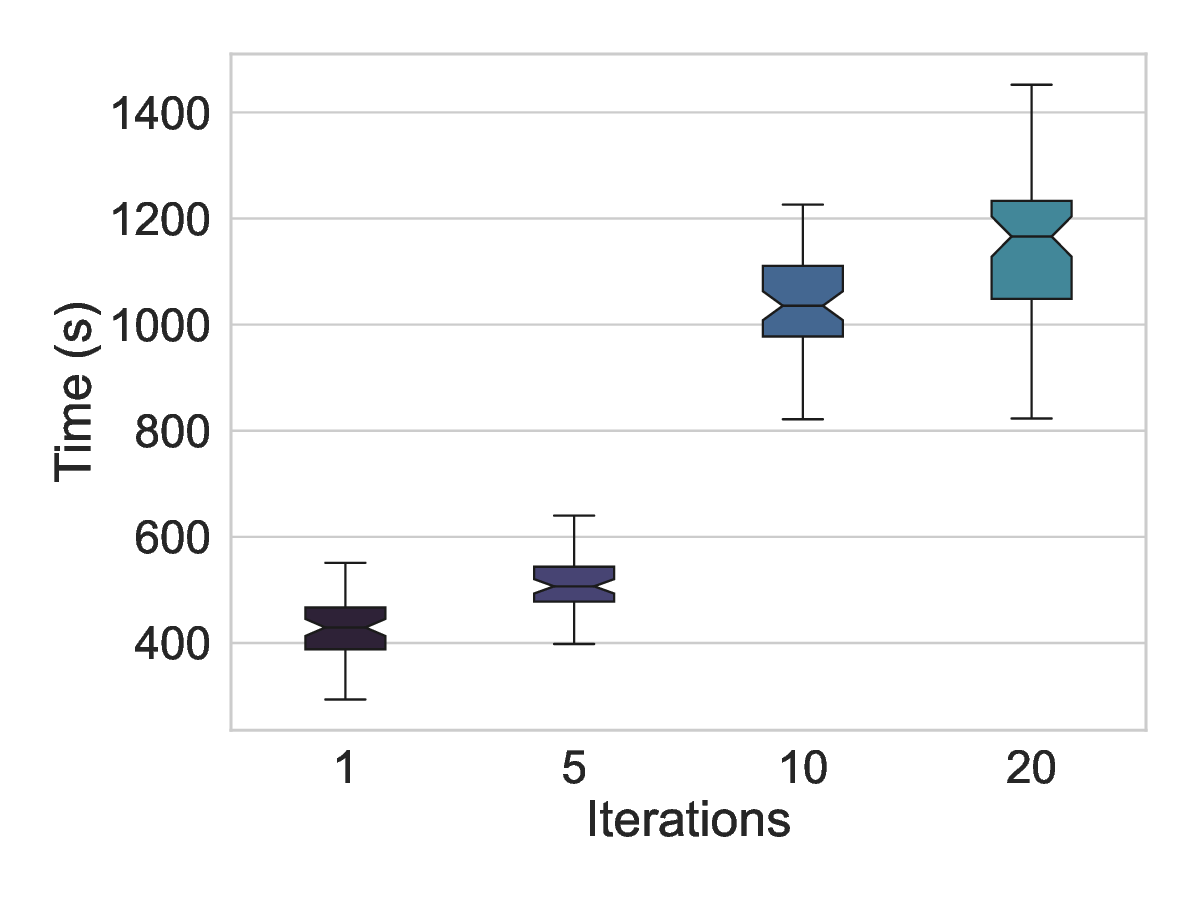}
        \caption{Distributions of \textbf{training time} for different numbers of perturbation iterations.}
        \label{fig: Training Time Different Iterations}
    \end{figure}

    As illustrated in \cref{Comparison on Training Time}, the perturbation process incurs additional computational costs. In this section, we investigate the impact of the number of perturbations on the final performance. We explored four different perturbation iterations, namely $\{1,5,10,20\}$. The statistical pairwise comparison of test $R^2$ scores for different numbers of perturbation iterations is presented in \cref{tab: Iteration of Perturbations}. The results indicate that using 10 iterations to estimate variance is sufficient to achieve good test $R^2$ scores, and its training time falls within acceptable ranges, as shown in \cref{fig: Training Time Different Iterations}. In the field of deep learning, it has been observed that increasing the number of iterations to solve the sharpness estimation problem could lead to changes in the optimal perturbation magnitude, but it did not result in improved test performance~\cite{andriushchenko2022towards}. In the neural network domain, Gaussian perturbations with approximately 3 to 10 iterations are adequate to yield good results~\cite{wang2021generalization}. This range changes to 10 to 20 in GP-based feature construction.

    \begin{table}[!htb]
        \caption{Statistical comparison of \textbf{test $R^2$ scores} using different numbers of perturbation iterations.}
        \label{tab: Iteration of Perturbations}
        \centering
        \begin{tabular}{cccc}%
            \toprule%
            & \textbf{5}              & \textbf{10}             & \textbf{20}             \\%
            \midrule%
            \textbf{1}  & 4(+)/33($\sim$)/21({-}) & 5(+)/27($\sim$)/26({-}) & 7(+)/25($\sim$)/26({-}) \\%
            \textbf{5}  & ---                     & 2(+)/47($\sim$)/9({-})  & 5(+)/43($\sim$)/10({-}) \\%
            \textbf{10} & ---                     & ---                     & 3(+)/54($\sim$)/1({-})  \\%
            \bottomrule%
        \end{tabular}%
    \end{table}

    \subsection{Analysis of the Magnitude of Perturbations}
    \def\ThreeFiveBetter{15}
    \def\ThreeFiveWorse{6}
    \def\ThreeOneBetter{24}
    \def\ThreeOneWorse{5}
    In the proposed method, the magnitude of perturbation is another hyperparameter. In this section, we conduct comparative experiments with standard deviations of $\{0.1, 0.3, 0.5\}$ to investigate the impact of the magnitude of perturbation on controlling overfitting. \cref{tab: Impact for the Magnitude of Perturbations} presents pairwise statistical comparisons of test $R^2$ scores regarding the magnitude of perturbations. From the experimental results, it is evident that selecting an appropriate magnitude for the perturbations is crucial for SAM. This finding is consistent with the application of SAM techniques in deep learning algorithms~\cite{andriushchenko2022towards}. Increasing the magnitude from 0.3 to 0.5 leads to a deterioration in results on \ThreeFiveBetter{} datasets and improves results on only \ThreeFiveWorse{} datasets. Similarly, decreasing the magnitude from 0.3 to 0.1 worsens outcomes on \ThreeOneBetter{} datasets and improves results on only \ThreeOneWorse{} datasets. Overall, the experimental results suggest that 0.3 is a suitable starting point if not using cross-validation to select the optimal magnitude of perturbation.

    \begin{table}[!htb]
        \caption{Statistical comparison of \textbf{test $R^2$ scores} using different magnitudes of perturbations.}
        \label{tab: Impact for the Magnitude of Perturbations}
        \centering
        \begin{tabular}{ccccc}%
            \toprule%
            & \textbf{0.1}            & \textbf{0.5}             \\%
            \midrule%
            \textbf{0.3} & 24(+)/29($\sim$)/5({-}) & 15(+)/37($\sim$)/6({-})  \\%
            \textbf{0.1} & ---                     & 13(+)/21($\sim$)/24({-}) \\%
            \bottomrule%
        \end{tabular}%
    \end{table}

    \section{Comparison of Ensemble Size}
    For the ensemble version of SAM-GP, the ensemble size is a parameter worth considering. This paper follows the convention of using 100 models in the final ensemble~\cite{zhang2021evolutionary}. In this section, we present the changes in test $R^2$ scores with varying ensemble sizes, as shown in \cref{tab: Ensemble Size}. The experimental results indicate that using an ensemble size of 100 yields the best results, but the difference between an ensemble size of 30 and 100 is not very significant. In fact, with the greedy ensemble selection technique~\cite{zhang2023sr}, it is possible to bridge the gap between an ensemble size of 30 and 100, which merits further exploration.

    \begin{table}[!htb]
        \caption{Statistical comparison of \textbf{test $R^2$ scores} using different ensemble sizes.}
        \label{tab: Ensemble Size}
        \centering
        \begin{tabular}{cccc}%
            \toprule%
            & \textbf{10}            & \textbf{30}            & \textbf{100}           \\%
            \midrule%
            \textbf{5}  & 0(+)/58($\sim$)/0({-}) & 0(+)/58($\sim$)/0({-}) & 0(+)/53($\sim$)/5({-}) \\%
            \textbf{10} & ---                    & 0(+)/58($\sim$)/0({-}) & 0(+)/56($\sim$)/2({-}) \\%
            \textbf{30} & ---                    & ---                    & 0(+)/58($\sim$)/0({-}) \\%
            \bottomrule%
        \end{tabular}%
    \end{table}

    \section{Further Analysis of Sharpness Estimation}
    In this paper, several default settings are directly drawn from the deep learning community, such as adaptive noise scaling~\cite{kwon2021asam} and the 1-sharpness estimation strategy~\cite{andriushchenko2022towards}. However, although they have a strong theoretical foundation, as shown in \cref{Theoretical Analysis}, they are not the only choices for sharpness estimation. This section provides a more in-depth analysis of these strategies.

    \subsection{Comparison of Noise Types}
    In this paper, we use Gaussian noise $\mathcal{N}(0,1)$ to perturb GP trees. Gaussian noise is chosen due to its simplicity for theoretical analysis and its effectiveness in controlling overfitting in neural networks. However, in practice, any type of noise can be used to perturb GP trees. In this section, we consider three types of noise:
    \begin{itemize}
        \item Uniform noise: Noise is sampled from a uniform distribution $\mathcal{U}(-1,1)$.
        \item Laplacian noise: Noise is sampled from a Laplacian distribution $\mathcal{L}(0,1)$. The probability density distribution $f(\epsilon)$ is given by:
        \begin{equation}
            f(\epsilon)=\frac{1}{2} \exp \left(-|\epsilon|\right).
        \end{equation}
        \item Ensemble noise: Ensemble noise consists of three types of noise: Gaussian noise, uniform noise, and Laplacian noise. In each round of perturbation, one of these three noises is randomly selected using a random number $\xi \in \mathcal{U}(0,1)$,
        \begin{equation}
            \operatorname{Noise} =
            \begin{cases}
                \mathcal{N}(0,1) & \text{if } 0 \leq \xi < \frac{1}{3} \\
                \mathcal{U}(-1,1) & \text{if } \frac{1}{3} \leq \xi < \frac{2}{3} \\
                \mathcal{L}(0,1) & \text{if } \frac{2}{3} \leq \xi \leq 1 \\
            \end{cases}
        \end{equation}
        Then, noise values are sampled from the corresponding distribution.
    \end{itemize}

    \def\UniformGaussianBetter{6}
    \def\UniformGaussianWorse{19}

    The comparison of $R^2$ scores for these three types of noise is presented in \cref{tab: Distribution}. The experimental results show that there is no significant difference between using Gaussian noise and ensemble noise to perturb inputs. When using Laplacian noise, there is a decrease in test $R^2$ scores, as seen in the results. As for the uniform distribution, it has significant advantages on \UniformGaussianBetter{} datasets and significantly worse performance on \UniformGaussianWorse{} datasets. Therefore, in the real world, it is recommended to use cross-validation to determine the noise type, but the most widely used Gaussian noise is a good default type of noise to perturb semantics.
    \begin{table}[!htp]
        \centering
        \caption{Comparison of \textbf{test $R^2$ scores} using different selection operators.}
        \label{tab: Distribution}
        \begin{tabular}{cccc}%
            \toprule%
            & \textbf{Uniform}        & \textbf{Laplace}         & \textbf{Ensemble}       \\%
            \midrule%
            \textbf{Normal}  & 19(+)/33($\sim$)/6({-}) & 10(+)/45($\sim$)/3({-})  & 3(+)/54($\sim$)/1({-})  \\%
            \textbf{Uniform} & ---                     & 11(+)/24($\sim$)/23({-}) & 8(+)/31($\sim$)/19({-}) \\%
            \textbf{Laplace} & ---                     & ---                      & 2(+)/49($\sim$)/7({-})  \\%
            \bottomrule%
        \end{tabular}%
    \end{table}

    \subsection{Comaprison of M-Sharpness}
    In this paper, 1-sharpness is utilized for calculating sharpness due to its widespread adoption in the deep learning community, and it offers better regularization for inducing sparse models~\cite{andriushchenko2022towards}. In practice, M can be assigned any arbitrary value or even set equal to the number of samples. Therefore, this section provides an empirical comparison and considers three settings:
    \begin{itemize}
        \item $M=4$: In this method, 4 samples are randomly grouped as a mini-batch, and the maximum sharpness is calculated separately for each mini-batch.
        \item $M=N$: This corresponds to the traditional n-sharpness approach. Sharpness is calculated on the entire batch, and the sharpness value in the iteration corresponding to the maximum sharpness is selected as the final sharpness. Formally, $S_{\text{n-SAM}}$ is defined as:
        \begin{equation}
            S_{\text{n-SAM}} =  \max_{k = 1, \dots, K} \frac{1}{N} \sum_{i = 1, \dots, N} (\tilde{Y_i^k} - Y_i)^2 - (\hat{Y_i} - Y_i)^2.
        \end{equation}
        \item Gaussian Model Perturbation (GMP)~\cite{wang2021generalization}: GMP uses Gaussian noise instead of the gradient method to perturb neural networks to estimate sharpness. With $K$ rounds of perturbations $Y^k$, where $k \in [1, K]$, GMP optimizes the mean perturbed loss instead of the maximum perturbed loss. Thus, the sharpness $S_\text{GMP}$ is defined as:
        \begin{equation}
            S_\text{GMP} = \frac{1}{N} \frac{1}{K} \sum_{i = 1, \dots, N}  \sum_{k = 1, \dots, K} (\tilde{Y_i^k} - Y_i)^2 - (\hat{Y_i} - Y_i)^2
        \end{equation}
    \end{itemize}
    \def\OneSAMNSAMBetter{22}
    \def\OneSAMNSAMWorse{5}
    \def\OneSAMGMPBetter{23}
    \def\OneSAMGMPWorse{4}
    The experimental results are presented in \cref{tab: M-Sharpness}. The experimental results demonstrate that using 1-sharpness yields the highest test $R^2$ scores, followed by 4-SAM. Importantly, 1-SAM significantly outperforms n-SAM on \OneSAMNSAMBetter{} datasets while underperforming on \OneSAMNSAMWorse{} datasets. Thus, in conclusion, 1-SAM is recommended in sharpness-aware minimization for evolutionary feature construction compared to n-SAM, which is consistent with the findings in deep learning~\cite{andriushchenko2022towards}. The experimental results about GMP are interesting, as 1-SAM significantly outperforms n-SAM on \OneSAMGMPBetter{} datasets while underperforming on \OneSAMGMPWorse{} datasets. These results demonstrate that in the situation of multi-round random perturbation for estimating sharpness, taking the maximum sharpness over multiple rounds is better than taking the average sharpness over these rounds. In the future, this approach is worth validating in neural networks.

    \begin{table}[!htp]
        \centering
        \caption{Comparison of \textbf{test $R^2$ scores} using different batch sizes for sharpness estimation.}
        \label{tab: M-Sharpness}
        \begin{tabular}{cccc}%
            \toprule%
            & \textbf{4-SAM}         & \textbf{n-SAM}          & \textbf{GMP}            \\%
            \midrule%
            \textbf{1-SAM} & 6(+)/50($\sim$)/2({-}) & 22(+)/31($\sim$)/5({-}) & 23(+)/31($\sim$)/4({-}) \\%
            \textbf{4-SAM} & ---                    & 17(+)/38($\sim$)/3({-}) & 19(+)/35($\sim$)/4({-}) \\%
            \textbf{n-SAM} & ---                    & ---                     & 2(+)/56($\sim$)/0({-})  \\%
            \bottomrule%
        \end{tabular}%
    \end{table}

    \subsection{Effectiveness of Adaptive Sharpness}
    Adaptive sharpness optimization, based on the scale of weights, was initially proposed in the context of deep learning and has demonstrated superior performance compared to standard sharpness optimization~\cite{kwon2021asam}. In this paper, we employ the adaptive sharpness strategy because it is supported by a solid theoretical foundation, as shown in \cref{Theoretical Analysis}. However, it is still possible to consider two alternative strategies:
    \begin{itemize}
        \item Batch Adaptive: This strategy scales the noise based on the standard deviation of a batch of inputs rather than scaling each input separately.
        \begin{equation}
            \tilde{\psi}_{a}(X) = \psi_{a}(X) + \mathcal{N}(0, \sigma_{\psi_{a}}(X))
        \end{equation}
        \item Without Adaptive (W/O): This strategy does not utilize adaptive techniques. In this case, random values are directly added to the inputs without any scaling. In other words, noise is synthesized by the following equation:
        \begin{equation}
            \tilde{\psi}_{a}(X) = \psi_{a}(X) + \mathcal{N}(0, \sigma^2)
        \end{equation}
    \end{itemize}
    The experimental results are presented in \cref{tab: Adaptive}. The results indicate that instance-wise adaptive sharpness estimation achieves significantly better performance compared to adaptive sharpness based on the standard deviation of the entire dataset or simply ignoring adaptiveness. Thus, it is recommended to use the instance-wise adaptive sharpness strategy in sharpness-aware minimization of evolutionary feature construction. It is worth noting that the data are standardized before training. In cases where data are not standardized, the adaptive sharpness strategy may be even more necessary to achieve good results.

    \begin{table}[!htp]
        \centering
        \caption{Comparison of \textbf{test $R^2$ scores} using different adaptive sharpness strategies.}
        \label{tab: Adaptive}
        \begin{tabular}{cccc}%
            \toprule%
            & \textbf{Batch}          & \textbf{W/O}            \\%
            \midrule%
            \textbf{Instance} & 12(+)/42($\sim$)/4({-}) & 12(+)/41($\sim$)/5({-}) \\%
            \textbf{Batch}    & ---                     & 1(+)/56($\sim$)/1({-})  \\%
            \bottomrule%
        \end{tabular}%
    \end{table}

    \begin{table*}[htbp]
        \centering
        \caption{The Hyperparameter Spaces of ML Methods used in Grid Search~\cite{cava2021contemporary}.}
        \begin{tabular}{l p{45em}}
            \toprule
            Method       & Hyperparameters                                                                                                                                                                                                                                                                                          \\
            \midrule
            AdaBoost     & \{`learning\_rate': (0.01, 0.1, 1.0, 10.0), `n\_estimators': (10, 100, 1000)\}                                                                                                                                                                                                                           \\
            KernelRidge  & \{`kernel': (`linear', `poly', `rbf', `sigmoid'), `alpha': (0.0001, 0.01, 0.1, 1), `gamma': (0.01, 0.1, 1, 10)\}                                                                                                                                                                                         \\
            LGBM         & \{`n\_estimators': (10, 50, 100, 250, 500, 1000), `learning\_rate': (0.0001, 0.01, 0.05, 0.1, 0.2), `subsample': (0.5, 0.75, 1), `boosting\_type': (`gbdt', `dart', `goss')\}                                                                                                                            \\
            MLP          & \{`activation': (`logistic', `tanh', `relu'), `solver': (`lbfgs', `adam', `sgd'), `learning\_rate': (`constant', `invscaling', `adaptive')\}                                                                                                                                                             \\
            RandomForest & \{`n\_estimators': (10, 100, 1000), `min\_weight\_fraction\_leaf': (0.0, 0.25, 0.5), `max\_features': (`sqrt', `log2', None)\}                                                                                                                                                                           \\
            LR (SGD)     & \{`alpha': (1e-06, 0.0001, 0.01, 1), `penalty': (`l2', `l1', `elasticnet')\}                                                                                                                                                                                                                             \\
            XGB          & \{`n\_estimators': (10, 50, 100, 250, 500, 1000), `learning\_rate': (0.0001, 0.01, 0.05, 0.1, 0.2), `gamma': (0, 0.1, 0.2, 0.3, 0.4), `subsample': (0.5, 0.75, 1)\}                                                                                                                                      \\
            Operon       & \{`population\_size': (500,), `pool\_size': (500,), `max\_length': (50,), `allowed\_symbols': (`add,mul,aq,constant,variable',), `local\_iterations': (5,), `offspring\_generator': (`basic',), `tournament\_size': (5,), `reinserter': (`keep-best',), `max\_evaluations': (500000,)\}                  \\
            & \{`population\_size': (500,), `pool\_size': (500,), `max\_length': (25,), `allowed\_symbols': (`add,mul,aq,exp,log,sin,tanh,constant,variable',), `local\_iterations': (5,), `offspring\_generator': (`basic',), `tournament\_size': (5,), `reinserter': (`keep-best',), `max\_evaluations': (500000,)\} \\
            & \{`population\_size': (500,), `pool\_size': (500,), `max\_length': (25,), `allowed\_symbols': (`add,mul,aq,constant,variable',), `local\_iterations': (5,), `offspring\_generator': (`basic',), `tournament\_size': (5,), `reinserter': (`keep-best',), `max\_evaluations': (500000,)\}                  \\
            & \{`population\_size': (100,), `pool\_size': (100,), `max\_length': (50,), `allowed\_symbols': (`add,mul,aq,constant,variable',), `local\_iterations': (5,), `offspring\_generator': (`basic',), `tournament\_size': (3,), `reinserter': (`keep-best',), `max\_evaluations': (500000,)\}                  \\
            & \{`population\_size': (100,), `pool\_size': (100,), `max\_length': (25,), `allowed\_symbols': (`add,mul,aq,exp,log,sin,tanh,constant,variable',), `local\_iterations': (5,), `offspring\_generator': (`basic',), `tournament\_size': (3,), `reinserter': (`keep-best',), `max\_evaluations': (500000,)\} \\
            & \{`population\_size': (100,), `pool\_size': (100,), `max\_length': (25,), `allowed\_symbols': (`add,mul,aq,constant,variable',), `local\_iterations': (5,), `offspring\_generator': (`basic',), `tournament\_size': (3,), `reinserter': (`keep-best',), `max\_evaluations': (500000,)\}                  \\
            FFX          & N/A                                                                                                                                                                                                                                                                                                      \\
            \bottomrule
        \end{tabular}
        \label{fig:SRBench ML-Hyperparameters}
    \end{table*}

    \onecolumn
    \setlength{\LTcapwidth}{\textwidth}
    \begin{longtable}[htbp]{lrrrrrrrr}
        \caption{Detailed mean test $R^2$ of top-performing algorithms on all 58 PMLB datasets. Some extremely poor results with values lower than -1 have been replaced with "NaN".} \\
        \label{tab: Detailed Results} \\
        \toprule
        & SAM-EGP   & SAM-GP    & AdaBoost  & RandomForest & XGB       & MLP       & LGBM      \\
        \midrule
        1027 & 0.854726  & 0.854561  & 0.787393  & 0.818687     & 0.766583  & 0.855499  & 0.797568  \\
        1028 & 0.308487  & 0.290274  & 0.291492  & 0.247797     & 0.072232  & 0.211904  & 0.173838  \\
        1029 & 0.532151  & 0.532176  & 0.416140  & 0.399828     & 0.292998  & 0.485666  & 0.351490  \\
        1030 & 0.342372  & 0.343076  & 0.218239  & 0.193312     & 0.029510  & 0.284981  & 0.195258  \\
        1089 & 0.710169  & 0.693212  & 0.663939  & 0.638205     & 0.494464  & 0.730050  & 0.490654  \\
        1096 & 0.782031  & 0.779123  & 0.654888  & 0.739644     & 0.631863  & 0.798300  & 0.590756  \\
        1191 & 0.168656  & 0.134398  & 0.180747  & 0.202281     & 0.042035  & -0.092548 & 0.065320  \\
        1193 & 0.523828  & 0.499217  & 0.507364  & 0.514330     & 0.396540  & 0.438033  & 0.451329  \\
        1196 & 0.406076  & 0.384779  & 0.344141  & 0.352092     & 0.191686  & 0.256396  & 0.264480  \\
        1199 & 0.390203  & 0.381523  & 0.318973  & 0.346348     & 0.227642  & 0.248171  & 0.210688  \\
        1201 & 0.004921  & -0.035013 & -0.118123 & -0.111686    & -0.360642 & -0.232617 & -0.005870 \\
        1203 & 0.503840  & 0.478036  & 0.442429  & 0.486452     & 0.355591  & 0.407642  & 0.387054  \\
        1595 & -0.031270 & -0.076824 & -0.108539 & -0.087267    & -0.316255 & -0.404611 & -0.004469 \\
        192  & 0.467214  & 0.401665  & 0.502347  & 0.505674     & 0.345272  & 0.560730  & 0.323541  \\
        195  & 0.782467  & 0.750412  & 0.823562  & 0.846289     & 0.789803  & 0.784240  & 0.763104  \\
        197  & 0.965190  & 0.956489  & 0.944252  & 0.938034     & 0.942572  & 0.527992  & 0.934283  \\
        201  & 0.750311  & 0.756006  & 0.760735  & 0.786726     & 0.749537  & 0.287320  & 0.736526  \\
        207  & 0.770959  & 0.749170  & 0.804841  & 0.842715     & 0.783954  & 0.794989  & 0.756793  \\
        210  & 0.820529  & 0.803585  & 0.735931  & 0.768603     & 0.768930  & 0.837155  & 0.669371  \\
        215  & 0.910450  & 0.889931  & 0.876456  & 0.893822     & 0.888945  & 0.840620  & 0.882717  \\
        218  & 0.431111  & 0.416437  & 0.407862  & 0.443194     & 0.352278  & 0.247238  & 0.271315  \\
        225  & 0.546320  & 0.504644  & 0.469742  & 0.492196     & 0.403452  & 0.186283  & 0.406376  \\
        227  & 0.952632  & 0.946648  & 0.932127  & 0.931276     & 0.931976  & 0.621185  & 0.921244  \\
        228  & 0.559855  & 0.488469  & 0.670801  & 0.671009     & 0.636026  & 0.678584  & 0.617867  \\
        229  & 0.841345  & 0.825342  & 0.795461  & 0.811418     & 0.800843  & 0.785734  & 0.783997  \\
        230  & 0.753547  & 0.790716  & 0.761560  & 0.772730     & 0.715084  & 0.794769  & 0.750850  \\
        294  & 0.766088  & 0.735205  & 0.742512  & 0.750332     & 0.662901  & 0.753038  & 0.704116  \\
        344  & 0.996610  & 0.995030  & 0.916482  & 0.937511     & 0.942911  & 0.924209  & 0.937431  \\
        4544 & 0.695669  & 0.673989  & 0.579617  & 0.583654     & 0.475654  & 0.380311  & 0.370879  \\
        485  & 0.466712  & 0.477391  & 0.461214  & 0.479891     & 0.355181  & 0.408427  & 0.375554  \\
        503  & 0.708665  & 0.705272  & 0.643500  & 0.664546     & 0.595600  & 0.695853  & 0.580727  \\
        505  & 0.996106  & 0.995856  & 0.979081  & 0.978621     & 0.982174  & 0.954379  & 0.970657  \\
        519  & 0.735646  & 0.722003  & 0.696015  & 0.651892     & 0.596941  & 0.732528  & 0.662674  \\
        522  & 0.161427  & 0.131922  & 0.190446  & 0.237460     & 0.095320  & 0.088440  & 0.087532  \\
        523  & 0.940811  & 0.938427  & 0.943045  & 0.945999     & 0.941825  & 0.936877  & 0.944799  \\
        527  & 0.952894  & 0.948503  & 0.717565  & 0.722000     & 0.672494  & NaN       & 0.541663  \\
        529  & 0.774732  & 0.776661  & 0.495957  & 0.534237     & 0.478548  & 0.751388  & 0.502199  \\
        537  & 0.621150  & 0.607407  & 0.484567  & 0.507297     & 0.379761  & 0.581172  & 0.388682  \\
        542  & 0.167115  & 0.192564  & 0.431409  & 0.357981     & 0.145668  & 0.164405  & 0.034206  \\
        547  & 0.469542  & 0.463897  & 0.475554  & 0.486000     & 0.402307  & 0.418915  & 0.411154  \\
        556  & 0.866024  & 0.857930  & 0.844026  & 0.863950     & 0.821196  & 0.809733  & 0.787475  \\
        557  & 0.868733  & 0.867873  & 0.844251  & 0.891474     & 0.855027  & 0.848002  & 0.802575  \\
        560  & 0.979913  & 0.982663  & 0.942899  & 0.968217     & 0.967661  & 0.947857  & 0.952834  \\
        561  & 0.929144  & 0.924079  & 0.707248  & 0.776173     & 0.853230  & 0.851298  & 0.876368  \\
        562  & 0.952632  & 0.946648  & 0.932127  & 0.931276     & 0.931976  & 0.621185  & 0.921244  \\
        564  & 0.911372  & 0.900269  & 0.666915  & 0.665891     & 0.626455  & 0.731343  & 0.673700  \\
        573  & 0.965190  & 0.956489  & 0.944252  & 0.938034     & 0.942572  & 0.527992  & 0.934283  \\
        574  & 0.211307  & 0.176390  & 0.318173  & 0.324906     & 0.182810  & -0.007378 & 0.110001  \\
        659  & 0.622348  & 0.592591  & 0.511390  & 0.529914     & 0.254964  & 0.564836  & 0.307717  \\
        663  & 0.995971  & 0.993546  & 0.952503  & 0.976445     & 0.985812  & 0.991243  & 0.973437  \\
        665  & 0.305412  & 0.310292  & 0.163178  & 0.208788     & -0.098119 & 0.205274  & 0.072138  \\
        666  & 0.563532  & 0.542018  & 0.491946  & 0.508282     & 0.420062  & 0.539845  & 0.339060  \\
        678  & 0.275103  & 0.273002  & 0.044757  & 0.111072     & -0.176710 & 0.185851  & 0.020555  \\
        687  & 0.475916  & 0.400928  & 0.428099  & 0.468948     & 0.298187  & 0.433818  & 0.234628  \\
        690  & 0.959882  & 0.956200  & 0.951978  & 0.959725     & 0.958619  & 0.933151  & 0.959247  \\
        695  & 0.859767  & 0.855277  & 0.824942  & 0.835866     & 0.780648  & 0.785195  & 0.773405  \\
        706  & 0.577552  & 0.502538  & 0.503907  & 0.586631     & 0.474260  & 0.552942  & 0.389710  \\
        712  & 0.739643  & 0.737850  & 0.704985  & 0.691104     & 0.622566  & 0.750884  & 0.668710  \\
        \bottomrule
    \end{longtable}

    \twocolumn

\end{document}